\documentclass[final,5p,times,twocolumn,authoryear]{elsarticle}

\usepackage[T1]{fontenc}
\usepackage[utf8]{inputenc}

\usepackage{graphicx}
\usepackage{comment}
\usepackage{amsmath}
\usepackage{amssymb}
\usepackage{textcomp}

\usepackage{tikz}
\usepackage{algorithm}
\usepackage{algpseudocode}

\usepackage{xcolor}
\usepackage{verbatim}
\usepackage[flushleft]{threeparttable}

\usepackage{array}
\usepackage{ragged2e}
\usepackage{booktabs}
\usepackage{tabularx}
\usepackage{colortbl}
\usepackage{multirow}
\usepackage{enumitem}

\usepackage{adjustbox}
\usepackage{blkarray}
\usepackage{bigstrut}

\usepackage{epstopdf}
\usepackage{float}
\usepackage{grffile}
\usepackage[hyphens]{url}

\newcolumntype{Y}{>{\RaggedRight\arraybackslash}X}
\newcolumntype{L}[1]{>{\RaggedRight\arraybackslash}p{#1}}

\usepackage[
    breaklinks=true,
    hidelinks,
    pdfauthor={
        Saviz Changizi,
        Nasibeh Mohammadzadeh,
        Mohammad Shojafar,
        and Rahim Tafazolli
    }
]{hyperref}

\newcommand{\codefield}[1]{\nolinkurl{#1}}
\definecolor{maroon}{cmyk}{0,0.87,0.68,0.32}

\journal{Journal of Network and Computer Applications}

\let\cite\citep


\begin{document}

\begin{frontmatter}

\title{
When Does AI Supervision Help? A Role-Aware Study of Network Fraud Decision Management with Blockchain Auditability
}

\author[inst1]{Saviz Changizi\corref{cor1}}
\ead{s.changizi@surrey.ac.uk}

\author[inst2]{Nasibeh Mohammadzadeh}
\ead{nasibeh.m@gmail.com}

\author[inst1]{Mohammad Shojafar}
\ead{m.shojafar@surrey.ac.uk}

\author[inst1]{Rahim Tafazolli}
\ead{r.tafazolli@surrey.ac.uk}

\cortext[cor1]{Corresponding author}

\address[inst1]{
6G Innovation Centre,
Institute for Communication Systems,
University of Surrey,
Guildford,
United Kingdom
}

\address[inst2]{
Department of Network Engineering,
Universitat Politècnica de Catalunya (UPC),
Barcelona,
Spain
}

\begin{abstract}
When does a second artificial intelligence (AI) component improve a primary network-fraud decision rather than add operational burden? Our results show that supervision is not inherently beneficial: under the evaluated role assignments and operating policies, additional fraud recovery is usually accompanied by enough legitimate-user blocking to worsen the overall error profile. We study this question through a role-aware Decider--Supervisor (DS) framework with blockchain auditability, evaluating four directional configurations that combine centralised machine learning, a Federated Averaging (FedAvg)-trained federated meta-model, and Base or Quantized Low-Rank Adaptation (QLoRA) large language model variants. The analysis compares primary-only and supervised decisions using non-hard fraud performance, intervention burden, paired uncertainty, conditional threshold calibration, traffic-mix and \textsc{Review}-capacity sensitivity, dependability tests, and blockchain lifecycle controls. The deterministic hard gate resolves 89.994\% of fraudulent requests, so the primary AI analysis focuses on the remaining non-hard population. Conditional validation calibration does not produce a consistently transferable supervisory advantage on deployment replay. Among the predefined supervised configurations, DS-3 QLoRA is the least disruptive, but it still underperforms its primary FedAvg stage in F1 and total errors. Across 36 reweighted traffic mixtures, supervision reduces total errors only for DS-4 Base in two extreme high-fraud scenarios. Blockchain tests support digest verification, tamper detection, authorisation, single-use review resolution, and post-finalisation integrity, while exposing a pre-finalisation single-write limitation. These findings show that the value of AI supervision depends on role assignment, conditional calibration, escalation policy, traffic composition, and lifecycle controls rather than on the presence of a second model alone.
\end{abstract}

\begin{keyword}
Network fraud decision management
\sep Decider--Supervisor framework
\sep trustworthy AI
\sep AI dependability
\sep federated learning
\sep large language models
\sep blockchain auditability
\end{keyword}

\end{frontmatter}

\section{Introduction}\label{sec:intro}

As telecommunication systems evolve towards the highly distributed architectures envisioned for 6G, networked services will increasingly depend on scalable, privacy-preserving, and trustworthy coordination across heterogeneous computing and communication domains \cite{valdezsolis2026privacy}. Telecom fraud remains difficult to detect because fraudsters may act cooperatively and conceal their activities within large populations of legitimate users, which limits the effectiveness of methods that assess users in isolation \cite{wu2024fraud}. Severe class imbalance in telecom-fraud datasets can bias learning towards the legitimate majority and reduce sensitivity to the comparatively rare fraud class \cite{hu2024gatcobo}.

Centralised and federated learning (FL) have both been applied to network-intrusion detection, with experimental comparisons revealing distinct performance and scalability profiles across the two approaches \cite{belenguer2023gowfed}. Federated intrusion-detection systems support collaborative model training without centralising raw local data, but non-independent and identically distributed (non-IID) data and privacy protection remain important deployment considerations \cite{singh2024flids}. Related privacy-preserving federated-learning research also highlights statistical and model heterogeneity and communication overheads across distributed clients \cite{han2026fedehp}. Large language models (LLMs) add language-generation and reasoning capabilities to such systems, but they can also produce plausible yet factually incorrect outputs with unwarranted confidence, raising concerns about uncertainty representation and reliability in consequential applications \cite{shorinwa2025uncertainty}.

Recent network-security research increasingly treats intrusion detection as one stage of a broader mitigation and response pipeline rather than as an isolated classification task \cite{mhamdi2024securing}. In 5G and beyond, intrusion-response selection must balance security effectiveness against quality-of-service impacts and the monetary, time, and energy costs of countermeasures \cite{bozorgchenani2024intrusion}. A related trade-off arises at the detector level, where the operating threshold affects both attack detection and false alarms and, consequently, the volume of traffic subjected to blocking or further intervention \cite{racherla2024deepids}. 

Trustworthy artificial intelligence (AI) should be assessed at the system level, where AI models operate as components of broader software systems whose trustworthiness must be considered across the software lifecycle \cite{liu2024gap}. Operationalising this view requires multidimensional evaluation strategies and quantitative metrics spanning the machine-learning lifecycle \cite{kemmerzell2025trust}. Research on human--AI decision making also shows that adding another decision maker does not necessarily improve outcomes: on average, human--AI systems performed worse than the better-performing standalone human or AI system, with particularly negative results in decision-making tasks \cite{vaccaro2024humanai}.

Blockchain is increasingly used in intrusion-detection systems as a decentralised trust and audit layer that supports tamper-evident recording, distributed coordination, and smart-contract-based policy enforcement \cite{chen2026blockchainids}. In secure FL, blockchain has also been combined with consensus-based model verification, transparent participation mechanisms, and personalised differential privacy to improve resilience against poisoning attacks while protecting client information \cite{wu2026blockchainfl}. Nevertheless, blockchain-integrated deep-learning intrusion detection in industrial-control settings still lacks mature end-to-end integration and comprehensive evaluation of overall system effectiveness \cite{devipriya2025blockchain}.

Centralised--federated comparisons in network-intrusion detection mainly examine detector performance and scalability \cite{belenguer2023gowfed}, while federated intrusion-detection studies also address training-stage issues such as non-IID data and privacy protection \cite{singh2024flids}. Work on trustworthy AI broadens evaluation to the software-system lifecycle \cite{liu2024gap}, and blockchain-integrated intrusion-detection research points to gaps in end-to-end integration and system-level evaluation \cite{devipriya2025blockchain}. Existing studies do not yet explain how heterogeneous AI components should cooperate after inference, how predefined role directions affect final decisions, or when additional fraud recovery justifies legitimate-request blocking and \textsc{Review} workload. We examine these questions through a unified evaluation of decision quality, intervention burden, dependability, and blockchain-supported lifecycle enforcement.

Our companion study introduced a blockchain-linked decision-management pipeline that transformed heterogeneous AI risk outputs into policy-resolved telecom/Internet of Things (IoT) fraud-control actions and recorded the resulting decision lifecycle \cite{changizi2026blockchainlinked}. The study treated its AI components as independent risk sources and did not evaluate them as cooperating components with asymmetric operational responsibilities. The present study reuses the centralised-ML, Federated Averaging (FedAvg)-trained federated meta-model, and LLM outputs and validation-derived thresholds from the companion pipeline without retraining the detectors. It then evaluates their downstream behaviour within the decision pipeline under explicitly assigned Decider and Supervisor roles.

The proposed framework evaluates four directional configurations under a conservative escalation policy, in which a Decider produces the primary approve/block outcome and a Supervisor reassesses a bounded subset of primary approvals. The evaluation covers incremental fraud recovery and legitimate-request burden, sensitivity to traffic composition and finite \textsc{Review} capacity, dependability under controlled faults and score manipulation, and blockchain-supported submission, resolution, finalisation, and authorisation controls. The main contributions of this paper are as follows.

\begin{itemize}[leftmargin=*]

\item We formulate network-fraud decision management as a role-aware Decider--Supervisor (DS) problem and evaluate four configurations, DS-1--DS-4, spanning centralised-ML--LLM and FedAvg--LLM pairings in both implemented role directions. This shifts evaluation from isolated detector performance to complete role-aware decision pipelines.

\item We define bounded supervisory intervention through explicit candidacy, evidence-validity, no-downgrade, threshold, and resolution constraints. Its effects are measured relative to the primary-only stage through incremental fraud recovery, false-positive burden, recall, legitimate-request blocking, total errors, \textsc{Review} workload, and block-rate changes.

\item We evaluate operational sensitivity across Base and Quantized Low-Rank Adaptation (QLoRA) LLM variants, binary and three-way \textsc{Review} interpretations, 36 exact traffic mixtures, and 11 finite \textsc{Review}-capacity levels. The analysis includes matched score-transfer auditing, label-free queue prioritisation, Pareto-frontier analysis, and normalised cost-ratio sensitivity. Oracle adjudication and cost ratios are used only as analytical bounds, not as deployable reviewer performance or empirically estimated operator costs.

\item We combine a formal threat-and-fault model with controlled dependability tests covering threshold and resolver sensitivity, unavailable Supervisor evidence, systematic and targeted score manipulation, LLM parsing and fallback faults, and alternative primary-component failure-resolution policies.

\item We integrate resolved decisions with a blockchain audit and enforcement layer and test digest reconstruction, payload tampering, conflicting or repeated submissions, \textsc{Review} resolution, finalisation, post-finalisation mutation, and endpoint authorisation. The evaluation reports both the enforced protections and the identified pre-finalisation single-write limitation without treating blockchain as a source of predictive performance.

\end{itemize}

The remainder of this paper is organised as follows. Section~\ref{sec:related} reviews related work and positions the research gap. Sections~\ref{sec:system_model} and \ref{sec:framework} define the system model and present the proposed framework. Section~\ref{sec:experimental_design} describes the experimental design, followed by the results in Section~\ref{sec:results} and their discussion in Section~\ref{sec:discussion}. Section~\ref{sec:threats} addresses threats to validity and limitations, and Section~\ref{sec:conclusion} concludes the paper.

\subsection{Relationship to Prior Work by the Authors}
\label{subsec:relationship_prior_work}

The present study builds on two earlier studies by the authors \citep{changizi2025hpcc,changizi2026blockchainlinked} but addresses a different research question. Table~\ref{tab:author_prior_work} makes the reuse boundary and the additional contribution explicit. Elements reused directly from the companion study, including the dataset, preprocessing pipeline, component outputs, validation-derived operating artefacts, and blockchain-linked components, are treated as fixed inputs rather than as new contributions of the present paper. The contribution examined here is the behaviour of those retained components when they are assigned asymmetric Decider and Supervisor responsibilities and evaluated through intervention burden, conditional operating behaviour, directional uncertainty, dependability, and blockchain-supported decision-lifecycle controls.

\begin{table*}[!t]
\centering
\caption{Relationship between the present study and prior work by the authors.}
\label{tab:author_prior_work}
\scriptsize
\renewcommand{\arraystretch}{1.18}
\setlength{\tabcolsep}{3pt}

\begin{tabularx}{\textwidth}{@{}L{2.2cm} Y Y Y Y@{}}
\toprule
\textbf{Study} &
\textbf{Research question} &
\textbf{Reused elements} &
\textbf{New elements in that study} &
\textbf{Principal conclusion} \\
\midrule

HPCC 2025~\citep{changizi2025hpcc}
&
How can heterogeneous ML, LLM, and blockchain components be combined within telecom-fraud analysis configurations?
&
Conceptual precedent for heterogeneous ML--LLM--blockchain integration and for distributing decision authority between ML and LLM components; no HPCC dataset or trained detector is directly reused in the present study.
&
Six heterogeneous configurations with different distributions of ML/LLM decision authority, including threshold-based override, consultation, and fallback behaviour.
&
Heterogeneous AI and blockchain components can be integrated within alternative fraud-analysis architectures, but supervisory effects relative to a common primary-only baseline were not isolated.
\\

\midrule

Companion study~\citep{changizi2026blockchainlinked}
&
How can heterogeneous AI risk outputs be converted into auditable operational telecom/IoT fraud-control decisions?
&
Dataset and preprocessing pipeline, standalone M1--M3 outputs, validation-derived operating artefacts, and blockchain-linked audit components reused directly by the present study.
&
Standalone M1--M3 risk-source artefacts, policy-resolved decision management, and a shared blockchain-linked audit and execution lifecycle.
&
Heterogeneous risk sources can be converted into policy-resolved actions and recorded through a common auditable lifecycle, without assigning the AI components asymmetric Decider--Supervisor roles.
\\

\midrule

\textbf{Present study}
&
\textbf{When does a second AI component improve or degrade a primary fraud decision under assigned Decider--Supervisor roles?}
&
\textbf{Dataset, preprocessing, standalone M1--M3 outputs, validation-derived baseline thresholds, and blockchain-linked audit components from the companion pipeline; no detector retraining.}
&
\textbf{Four directional DS configurations; primary-versus-supervised intervention analysis; non-hard population analysis; conditional-validation calibration sensitivity; paired directional uncertainty; Review-capacity and traffic-mix analyses; controlled fault and score-manipulation tests; and blockchain lifecycle-security assessment.}
&
\textbf{Supervision is not inherently beneficial: its value depends on role assignment, conditional threshold behaviour, intervention burden, resolver policy, traffic composition, and lifecycle controls.}
\\

\bottomrule
\end{tabularx}
\end{table*}

\section{Related Work}\label{sec:related}

Recent telecom-fraud work has focused on two recurring problems: relationships among users that are missed by isolated analysis and severe class imbalance. Latent synergy graph learning models cooperative fraudsters who disguise their activities within large populations of benign telecom users \cite{wu2024fraud}. Cost-sensitive graph attention has also been used to reduce majority-class bias and improve the classification of minority fraud cases \cite{hu2024gatcobo}. In both cases, the evaluation is centred on the predictive performance of a standalone detector.

Federated learning offers an alternative to centralised network-intrusion detection when traffic data are distributed across organisations or devices. A direct comparison of centralised and FedAvg-based intrusion-detection models examined detection performance and scalability under the same experimental setting \cite{belenguer2023gowfed}. Federated intrusion detection for next-generation networks has also been evaluated under privacy, class-imbalance, and non-IID data constraints \cite{singh2024flids}. Work on personalised FL further addresses statistical and model heterogeneity, privacy protection, and communication bottlenecks across distributed clients \cite{han2026fedehp}. FL has also been applied to transaction-fraud detection through a joint horizontal and vertical framework designed to improve fraud detection while addressing privacy protection and communication costs during collaborative training \cite{li2026privacyfl}. These studies focus on model training, distributed optimisation, privacy, heterogeneity, and detector performance.

Some network-security work moves beyond detection to consider response and mitigation. One study embedded a hybrid deep-autoencoder--random-forest detector in a three-layer software-defined networking (SDN) framework for intrusion prevention and attack mitigation \cite{mhamdi2024securing}. The evaluation included false-positive rate (FPR), controller throughput, and latency, covering both detection performance and controller overhead. Another study framed countermeasure selection in 5G and beyond as a trade-off between security and quality of service (QoS) \cite{bozorgchenani2024intrusion}. The selection process considered security benefit together with monetary cost, preparation and deployment time, and energy consumption. These studies extend evaluation beyond classification accuracy to mitigation effectiveness, system overhead, and response cost.

Threshold selection provides another link between detector output and operational action. In real-time IoT intrusion detection, the operating threshold affects both attack detection and false alarms and determines how much traffic is classified as non-benign and blocked \cite{racherla2024deepids}. This operational concern is consistent with a software-engineering view of trustworthy AI in which the model is evaluated as part of a wider AI-enabled system rather than in isolation \cite{liu2024gap}. Work on trustworthy AI also treats explainability as one contributor to trustworthiness alongside fairness, robustness, interpretability, and related properties \cite{ali2023xai}. Trustworthiness evaluation therefore needs to cover multiple dimensions and use quantitative, domain-appropriate metrics across the machine-learning lifecycle \cite{kemmerzell2025trust}. System quality depends on both model accuracy and the policy that converts model outputs into operational decisions.

Adding a second decision maker cannot be assumed to improve an initial decision. Research on human--AI interaction in safety-critical industries distinguishes the roles assigned to AI-enabled systems and identifies AI output, usage context, and measurement choices as central dimensions for characterising and evaluating the interaction \cite{bach2024humanai}. Human--AI combinations have, on average, performed worse than the better-performing standalone human or AI system, with particularly negative results in decision-making tasks \cite{vaccaro2024humanai}. Human--AI collaboration is not equivalent to cooperation between two AI components, so these findings do not predict how an AI Supervisor will behave. Even so, the evidence cautions against assuming that a second decision maker will improve decision quality. The value of secondary review needs to be assessed through the decisions changed, the additional correct interventions achieved, and the new errors introduced.

Related work on selective prediction, learning to defer, and classifier cascades provides a formal perspective on when a second decision stage should be invoked. Learning-to-defer formulations allow a model to pass selected decisions to a downstream decision maker while accounting for the downstream agent's performance \cite{madras2018defer}, and subsequent work jointly models prediction and deferral through a classifier and rejector under a cost-sensitive formulation \cite{mozannar2020defer}. Classification with a reject option similarly treats abstention as a cost-sensitive mechanism for avoiding risky misclassification \cite{charoenphakdee2021rejection}. In classifier cascades, confidence-based deferral can also be suboptimal when the downstream model has input-specific strengths or when the operating distribution changes \cite{jitkrittum2023cascade}. These approaches motivate selective rather than unconditional secondary intervention. The present study differs in that it does not learn a deferral policy or optimise a new classifier--rejector pair. Instead, it evaluates predefined AI--AI Decider--Supervisor configurations in which candidacy, evidence validity, operating thresholds, and resolution rules determine whether an existing secondary component can alter an existing primary decision.

Secure federated-learning work mainly focuses on protecting distributed model training and aggregation. Verifiable differential privacy has been combined with Byzantine-resilient FL through secondary servers that validate model updates before verifiable differential-privacy noise is added during aggregation \cite{zhang2025verifiablefl}. Model-contrastive FL has also addressed poisoning attacks, lightweight privacy preservation, and non-IID data heterogeneity within local optimisation and secure aggregation \cite{zhang2026contrastivefl}.

Blockchain-based FL has been combined with a proof-of-verification consensus protocol, personalised differential privacy, and a transparent reward mechanism \cite{wu2026blockchainfl}. In federated intrusion detection, blockchain-based trust management, smart-contract verification, and trust-weighted aggregation have been used to support secure collaborative model construction \cite{kamran2026blockchainfl}. A separate trust-aware federated intrusion-detection system combines differential-privacy evaluation, Fast Gradient Sign Method (FGSM)-based robustness assessment, trust-weighted aggregation, and a monitoring-only trust ledger that records client participation and trust scores throughout training \cite{alshammari2026securetrust}. These approaches focus on local updates, client participation, privacy, aggregation, and global-model construction. They do not examine how trained-model scores or classifications are subsequently reviewed and converted into operational decisions.

Blockchain-based intrusion-detection research covers a wider range of integrity, provenance, and coordination functions. A recent survey organises this literature into trusted-rule, machine-learning, deep-learning, and federated-learning paradigms \cite{chen2026blockchainids}. Across these settings, blockchain supports tamper-evident records, provenance for model and training artefacts, decentralised coordination, and smart-contract-based enforcement of encoded rules and policies. A review of blockchain-integrated deep-learning intrusion detection in industrial-control systems reports limited end-to-end integration, insufficient assessment of overall system effectiveness, and unresolved issues involving scalability, latency, interoperability, computational complexity, and resource consumption \cite{devipriya2025blockchain}. These capabilities strengthen integrity and auditability, but they do not by themselves establish that a recorded decision was accurate, proportionate, or operationally appropriate.

Our earlier study combined machine learning, LLMs, and blockchain within a heterogeneous telecom-fraud framework \cite{changizi2025hpcc}. It showed that these technologies could operate within the same fraud-analysis architecture and evaluated six configurations with different distributions of decision authority between ML and LLM components. Some configurations already used threshold-based override, consultation, or fallback logic. However, the study did not formalise these interactions as a unified DS framework, isolate the effect of supervision against a primary-only baseline, or measure the intervention burden and new errors introduced by supervisory changes.

The companion study moved from heterogeneous risk generation to blockchain-linked decision management for telecom/IoT fraud-control requests \cite{changizi2026blockchainlinked}. Alternative AI risk-source configurations were converted into policy-resolved operational actions, and the request-level lifecycle was recorded through a shared blockchain-linked audit and execution layer. The framework separated risk generation, policy resolution, and lifecycle recording, but the AI configurations remained alternative risk sources within that process. They were not evaluated as Decider and Supervisor components with asymmetric responsibilities. The study did not compare the implemented role directions or isolate the additional fraud recovery, false positives, legitimate-request blocking, and \textsc{Review} burden produced by selective supervisory intervention.

The studies reviewed above address detector performance and distributed training \cite{belenguer2023gowfed,singh2024flids}, operational response and system-level trustworthiness \cite{bozorgchenani2024intrusion,liu2024gap,vaccaro2024humanai}, and blockchain-supported auditability \cite{devipriya2025blockchain,changizi2026blockchainlinked} through separate evaluation settings. Even the closest prior architectures did not isolate supervisory effects against a primary-only baseline or evaluate directional role configurations together with intervention burden, finite \textsc{Review} capacity, dependability testing, and blockchain-supported lifecycle enforcement \cite{changizi2025hpcc,changizi2026blockchainlinked}.

The remaining question is how final decisions vary when a centralised-ML or FedAvg-based component is paired with an LLM in assigned DS roles, when selective supervision improves the primary decision, and whether any additional fraud recovery justifies the accompanying false positives, legitimate-request blocking, and \textsc{Review} workload. These effects also need to be examined across traffic compositions and finite \textsc{Review} capacities, under controlled faults and score manipulation, and alongside blockchain-supported controls over the downstream decision lifecycle. The framework presented in this paper evaluates centralised-ML--LLM and FedAvg--LLM configurations in both implemented role directions under a shared conservative escalation policy. The focus is on role assignment, intervention effects, operational sensitivity, dependability, and blockchain-supported lifecycle enforcement; training another standalone fraud detector is outside the study scope. While Table~\ref{tab:author_prior_work} clarifies the reuse boundary and contribution relative to the authors' own prior studies, Table~\ref{tab:comparison} positions the proposed framework against the broader set of representative studies reviewed above.

\begin{table*}[!t]
\caption{Comparison of representative studies with the proposed role-aware DS framework.}
\label{tab:comparison}
\centering
\scriptsize
\renewcommand{\arraystretch}{1.18}
\setlength{\tabcolsep}{3.2pt}

\begin{tabularx}{\textwidth}{
    p{2.35cm}
    X
    X
    X
    >{\centering\arraybackslash}p{1.45cm}
    >{\centering\arraybackslash}p{1.65cm}
    >{\centering\arraybackslash}p{1.55cm}}
\hline

\textbf{Study} &
\textbf{Primary focus} &
\textbf{AI setting} &
\textbf{Evaluation stage} &
\shortstack{\textbf{Post-inference}\\\textbf{cooperation}} &
\shortstack{\textbf{Intervention /}\\\textbf{fault analysis}} &
\shortstack{\textbf{Decision-lifecycle}\\\textbf{blockchain support}} \\
\hline

\citet{wu2024fraud}
&
Coordinated telecom-fraud detection
&
Latent-synergy graph learning
&
Detector evaluation
&
--
&
--
&
--
\\

\citet{hu2024gatcobo}
&
Imbalanced telecom-fraud detection
&
Cost-sensitive graph attention
&
Detector evaluation
&
--
&
--
&
--
\\

\citet{belenguer2023gowfed}
&
Federated network-intrusion detection
&
Centralised ML and FedAvg-based FL
&
Training and detector evaluation
&
--
&
--
&
--
\\

\citet{singh2024flids}
&
Next-generation federated intrusion detection
&
Collaborative FL across distributed clients
&
Training and detector evaluation
&
--
&
--
&
--
\\

\citet{mhamdi2024securing}
&
SDN intrusion detection and mitigation
&
Hybrid autoencoder--random forest
&
Detection-to-mitigation pipeline
&
--
&
$\triangle$
&
--
\\

\citet{bozorgchenani2024intrusion}
&
5G intrusion-response selection
&
Security--QoS response optimisation
&
Post-detection response
&
--
&
$\triangle$
&
--
\\

\citet{wu2026blockchainfl}
&
Secure blockchain-based FL
&
FL, blockchain consensus, and differential privacy
&
Training and model aggregation
&
--
&
$\triangle$
&
$\triangle$
\\

\citet{changizi2025hpcc}
&
Heterogeneous telecom-fraud detection
&
ML, LLM, and blockchain
&
Post-inference configuration and architecture evaluation
&
$\triangle$
&
$\triangle$
&
$\triangle$
\\

Companion work~\citep{changizi2026blockchainlinked}
&
Auditable telecom/IoT decision management
&
Alternative heterogeneous risk-source configurations and blockchain
&
Policy-resolved post-inference decision management
&
--
&
--
&
$\checkmark$
\\

\textbf{Proposed framework}
&
\textbf{Role-aware network-fraud decision management}
&
\textbf{Centralised-ML--LLM and FedAvg--LLM pairings}
&
\textbf{Role-aware post-inference decision resolution}
&
$\checkmark$
&
$\checkmark$
&
$\checkmark$
\\

\hline
\multicolumn{7}{l}{
\footnotesize
$\checkmark$: explicitly evaluated;
$\triangle$: partially or adjacently addressed;
--: not addressed within the study scope.
}
\end{tabularx}
\end{table*}

\section{System Model and Problem Definition}
\label{sec:system_model}

This section defines the network-fraud decision setting used in this study. The aim is to compare fraud-decision behaviour across predefined role assignments involving machine-learning, federated-learning, and large-language-model components under a conservative supervisory policy. The analysis focuses on predictive performance and on the measured error burden introduced by supervision, particularly false-positive inflation and legitimate-user blocking.

\subsection{Network-Fraud Decision Setting}

The system receives a stream of network requests. Each request is represented by a feature vector and associated metadata. For request $i$, the input instance is denoted by $x_i$, and the system produces a binary operational decision:
\[
a_i \in \{\mathrm{APPROVE}, \mathrm{BLOCK}\}.
\]

The ground-truth label is denoted by $y_i$, where $y_i=1$ represents a fraudulent request and $y_i=0$ represents a legitimate request. A correct block of a fraudulent request is counted as a true positive. An incorrect block of a legitimate request is counted as a false positive. In this setting, false negatives leave fraud unblocked, while false positives constitute a legitimate-user blocking burden by denying legitimate requests.

The decision problem involves a trade-off between fraud detection and legitimate-user preservation. A model may achieve high recall by blocking many requests, but such behaviour can be undesirable if it produces excessive legitimate-user blocking. For that reason, the evaluation uses fraud-oriented metrics, such as recall and F1 score, together with operational metrics, such as FPR and block rate.

\subsection{Decider--Supervisor Role Model}

Each configuration consists of two role-specific components: a Decider and a Supervisor. The Decider produces the primary operational action, while the Supervisor is permitted to evaluate only eligible primary approvals. Primary blocks are protected by a no-downgrade rule and are therefore not reconsidered by the Supervisor.

For request $i$, let $h_i\in\{0,1\}$ denote the shared hard-fraud gate, where $h_i=1$ represents an explicit hard-fraud condition. For requests not locked by the hard-fraud gate, the Decider produces the primary action
\[
a_i^{\mathrm{P}}\in\{\mathrm{APPROVE},\mathrm{BLOCK}\}.
\]

A primary approval becomes a candidate for supervisory evaluation only when it also satisfies the eligibility conditions of the selected role configuration. Let $\eta_i\in\{0,1\}$ denote the role-specific eligibility indicator. The supervisory candidate flag is defined as
\begin{equation}
\gamma_i =
\mathbb{1}\{h_i=0\}
\mathbb{1}\{a_i^{\mathrm{P}}=\mathrm{APPROVE}\}
\eta_i.
\label{eq:supervisory_candidate}
\end{equation}

The indicator $\eta_i$ represents the configuration-specific candidacy contract derived from the aligned role artefacts; no single universal feature rule is used. In DS-1, it requires LLM eligibility for a non-hard centralised-ML approval; in DS-2, no additional secondary eligibility condition is imposed beyond a non-hard LLM approval; in DS-3, it requires LLM eligibility for a valid non-hard FedAvg approval; and in DS-4, it additionally requires admissible LLM-primary evidence, including successful parsing and absence of fallback output, together with the FedAvg supervisor-candidate artefact. Consequently, $\eta_i$ may depend on role-specific artefact metadata as well as the selected configuration and LLM variant.

For a supervisory candidate, let $e_i\in\{0,1\}$ denote the configuration-specific evidence-availability indicator exported by the aligned secondary-component artefact. This indicator represents more than the physical presence of an output. Depending on the selected configuration, it may also incorporate component eligibility, successful LLM parsing, absence of fallback output, Phase-1 evidence availability, and valid probability and confidence ranges, where applicable.

Let $r_i\in\{0,1\}$ denote the policy-stage evidence-validity condition. For LLM Supervisors, this condition rechecks successful parsing, absence of fallback output, and finite risk and confidence values. For the centralised-ML Supervisor in DS-2, it requires a finite probability value; the probability-implied confidence is retained as audit metadata and is not used as an activation condition. For the FedAvg Supervisor in DS-4, finite probability and confidence values are required. These policy-stage checks are applied defensively even when part of the same validation has already been performed while constructing the Phase-2 availability artefact. The Supervisor activation indicator is
\begin{equation}
u_i=\gamma_i e_i r_i.
\label{eq:supervisor_activation}
\end{equation}

In the exported artefacts, both the Supervisor-reliability state and the Supervisor-activation state encode $u_i$. The condition $r_i$ captures the component-specific validity checks supported by the parse, fallback, probability, and confidence information, but is not retained as a separate lifecycle state.

When activated, the Supervisor provides a risk score $p_i^{\mathrm{S}}\in[0,1]$. Let $\tau_{\mathrm{R}}^{\mathrm{S}}$ and $\tau_{\mathrm{B}}^{\mathrm{S}}$ denote the baseline-derived review and block thresholds, respectively, with $\tau_{\mathrm{R}}^{\mathrm{S}}\leq\tau_{\mathrm{B}}^{\mathrm{S}}$. The supervisory outcome is
\[
z_i^{\mathrm{S}}\in
\{\mathrm{NI},\mathrm{RETAIN},\mathrm{REVIEW},\mathrm{BLOCK}\},
\]
where $\mathrm{NI}$ denotes the \textsc{Not Invoked} state. The outcome is defined as
\begin{equation}
z_i^{\mathrm{S}} =
\begin{cases}
\mathrm{NI},
& \gamma_i=0, \\[2pt]
\mathrm{RETAIN},
& \gamma_i=1 \text{ and } u_i=0, \\[2pt]
\mathrm{RETAIN},
& u_i=1 \text{ and } p_i^{\mathrm{S}}<\tau_{\mathrm{R}}^{\mathrm{S}}, \\[2pt]
\mathrm{REVIEW},
& u_i=1 \text{ and }
\tau_{\mathrm{R}}^{\mathrm{S}}\leq p_i^{\mathrm{S}}<\tau_{\mathrm{B}}^{\mathrm{S}}, \\[2pt]
\mathrm{BLOCK},
& u_i=1 \text{ and } p_i^{\mathrm{S}}\geq\tau_{\mathrm{B}}^{\mathrm{S}}.
\end{cases}
\label{eq:supervisory_outcome}
\end{equation}

Thus, unavailable or unreliable supervisory evidence cannot alter the primary approval. Reliable evidence below the review threshold also retains the primary decision, whereas reliable evidence in the review or block region produces the corresponding escalation outcome. This role-aware formulation allows the same component family to be evaluated as a Decider in one configuration and as a Supervisor in another, while preserving a common supervisory lifecycle across the four configurations.

\subsection{Conservative Binary Decision Resolution}

The framework applies a conservative no-downgrade resolution policy. A primary block, including a block produced by the shared hard-fraud gate, cannot be converted into an approval by the Supervisor. For primary approvals, the final action depends on whether the Supervisor is invoked and, when activated, whether its reliable evidence produces a \textsc{Retain}, \textsc{Review}, or \textsc{Block} outcome.

The implemented blockchain enforcement backend requires a binary final action and does not include a separate human-review outcome. Consequently, \textsc{Review} and \textsc{Block} remain distinct supervisory lifecycle states and are recorded separately in the audit metadata, but both are conservatively resolved as final blocks. By contrast, a \textsc{Not Invoked} or \textsc{Retain} outcome preserves the primary approval. For compact notation, let $\mathcal{E}_{\mathrm{S}}=\{\mathrm{REVIEW},\mathrm{BLOCK}\}$ denote the supervisory escalation outcomes and let $\mathcal{K}_{\mathrm{S}}=\{\mathrm{NI},\mathrm{RETAIN}\}$ denote the outcomes that preserve the primary approval. The final action is then defined as

\begin{equation}
a_i^{\mathrm{F}} =
\begin{cases}
\mathrm{BLOCK},
& a_i^{\mathrm{P}}=\mathrm{BLOCK}, \\[2pt]
\mathrm{BLOCK},
& a_i^{\mathrm{P}}=\mathrm{APPROVE},\
  z_i^{\mathrm{S}}\in\mathcal{E}_{\mathrm{S}}, \\[2pt]
\mathrm{APPROVE},
& a_i^{\mathrm{P}}=\mathrm{APPROVE},\
  z_i^{\mathrm{S}}\in\mathcal{K}_{\mathrm{S}}.
\end{cases}
\label{eq:final_action}
\end{equation}

The \textsc{Review} outcome is an audit-level escalation state, not a third final enforcement action. In a deployment containing a human-review or delayed-verification layer, this outcome could be resolved separately. In the present implementation, however, it is mapped to \textsc{Block} to preserve conservative fraud-control behaviour within the binary execution pathway.

A decision-changing supervisory intervention occurs when the Decider initially approves a request but the final resolver converts that approval into a block. The row-level intervention indicator is defined as

\begin{equation}
\iota_i =
\mathbb{1}
\left\{
a_i^{\mathrm{P}}=\mathrm{APPROVE}
\land
a_i^{\mathrm{F}}=\mathrm{BLOCK}
\right\}.
\label{eq:intervention}
\end{equation}

Review-mediated and direct-block interventions are retained as separate lifecycle indicators:

\begin{equation}
\iota_i^{\mathrm{R}} =
\mathbb{1}
\left\{
z_i^{\mathrm{S}}=\mathrm{REVIEW}
\right\},
\qquad
\iota_i^{\mathrm{B}} =
\mathbb{1}
\left\{
z_i^{\mathrm{S}}=\mathrm{BLOCK}
\right\}.
\label{eq:intervention_types}
\end{equation}

Under the current binary resolution policy, both escalation types change an eligible primary approval into a final block:

\[
\iota_i=\iota_i^{\mathrm{R}}+\iota_i^{\mathrm{B}}.
\]

This separation preserves the distinction between review-mediated blocking and direct supervisory blocking in the row-level lifecycle artefacts. Because both escalation outcomes become final blocks under the implemented binary resolver, the compact paper analysis reports their sum as the overall decision-changing intervention count. The separate indicators remain available for lifecycle inspection and audit, while the aggregate intervention measure captures their combined operational effect on fraud recall, legitimate false positives, and final block rate.

\subsection{Formal Threat and Fault Model}
\label{subsec:threat_fault_model}

The dependability evaluation distinguishes non-adversarial component faults from adversarial attempts to alter the decision or audit lifecycle. The protected assets include the primary decision, Supervisor evidence and score, supervisory candidate and intervention states, the resolved final action, the review-resolution state, exported decision payloads and digests, and the corresponding blockchain lifecycle records. The principal security and dependability properties are preservation of existing primary blocks, containment of unavailable or invalid supervisory evidence, deterministic resolution, authorised state modification, single-use review resolution, detection of exported-payload modification, lifecycle traceability, and protection of evaluated states after finalisation.

The threat model considers request-side input manipulation, faulty or compromised primary and supervisory components, systematic and targeted Supervisor-score manipulation, authorised-but-faulty validators, unauthorised smart-contract callers, off-chain artefact modification, and faulty decision-resolution logic. The principal trust boundaries separate request submission, primary and secondary component outputs, policy-based resolution, exported audit artefacts, and blockchain-maintained request state. The evaluation assumes that the local blockchain and administrative account are not compromised, authorised-validator private keys are not stolen, the retained artefacts are accessible for offline evaluation, and ground-truth labels are used only for experimental assessment. Consensus-layer attacks, host compromise, denial-of-service, model-training privacy attacks, side-channel leakage, and validator-key theft are outside the scope of the present study.

Each considered fault or threat is mapped to a controlled empirical assessment. These assessments include Supervisor-evidence unavailability, systematic score bias, targeted threshold-crossing manipulation, LLM parsing and fallback faults, primary-LLM output failure, exported-payload tampering, duplicate and conflicting decision submissions, repeated review resolution, post-finalisation mutation, and unauthorised state-changing contract calls. Support for a security or dependability property is limited to the tested component, endpoint, lifecycle state, and attacker-capability boundary. Any violation found in these tests is reported as a system limitation and is not obscured by aggregate fraud-detection performance.

\section{Proposed Role-Aware Decider--Supervisor Framework}
\label{sec:framework}

This section introduces the role-aware DS framework used in this study. The framework evaluates specified pairings of heterogeneous AI components, where one component produces the primary fraud decision and another component reviews selected risky approvals. The aim is to compare final fraud-decision behaviour, the measured error and legitimate-user blocking burden introduced by supervisory intervention, and blockchain-audited enforcement outcomes across the four implemented role-aware configurations.

The framework is not a dynamic model-selection or adaptive orchestration system. It evaluates four role-aware configurations under a shared conservative resolution policy. This structure keeps the comparison focused on role assignment, supervisor type, and LLM variant. Routing and run-time model selection are outside the comparison.

\subsection{Framework Overview}

\begin{figure*}[!t]
    \centering
    \vspace*{-1.0cm}
    \includegraphics[width=0.95\textwidth]{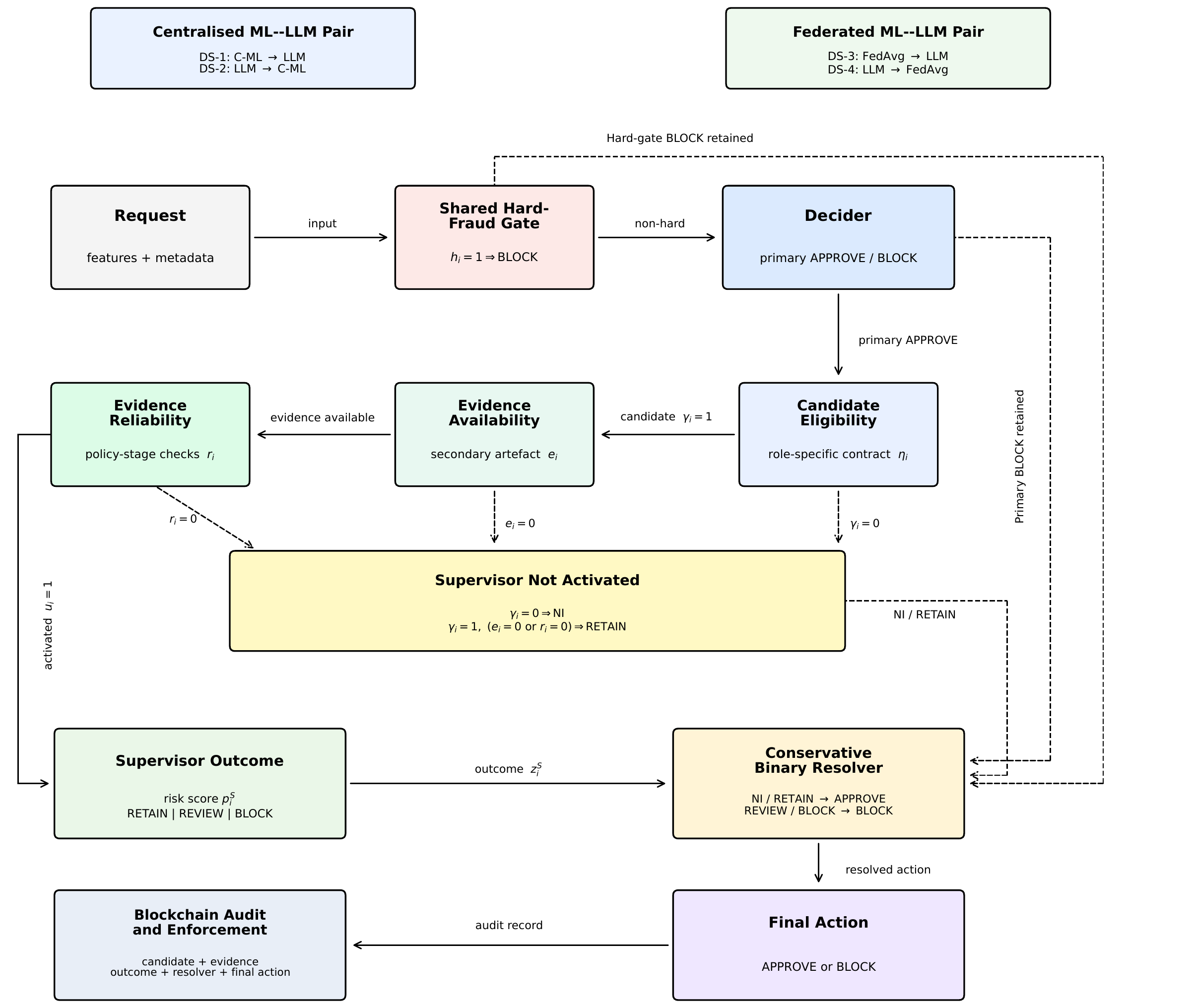}
    \caption{Role-aware DS decision lifecycle. Solid arrows show the active non-hard primary-approval pathway, while dashed arrows show protected hard-gate and primary-block bypasses and non-activation branches. Only eligible primary approvals become supervisory candidates. An ineligible primary approval produces a \textsc{Not Invoked} state, whereas unavailable or unreliable supervisory evidence produces \textsc{Retain}. When supervision is activated, the secondary risk score is mapped to \textsc{Retain}, \textsc{Review}, or \textsc{Block}. For primary approvals, the conservative binary resolver preserves \textsc{Not Invoked} and \textsc{Retain} outcomes as approvals and maps both \textsc{Review} and \textsc{Block} outcomes to final blocks before blockchain audit and enforcement.}
    \label{fig:framework}
\end{figure*}

Fig.~\ref{fig:framework} presents the architecture that implements the formal decision model defined in Section~\ref{sec:system_model}. It connects the shared hard-fraud gate, the assigned DS roles, role-specific candidacy and evidence checks, the conservative resolver, and the blockchain audit and enforcement layer. The figure distinguishes protected hard-gate and primary blocks from the supervisory pathway applied to eligible primary approvals. The configuration-specific role assignments are summarised in Table~\ref{tab:architecture_definitions}, while the lifecycle variables and resolution rules follow the definitions given in Section~\ref{sec:system_model}.

The framework separates fraud-decision logic from audit recording. The Decider and Supervisor determine the final approve/block outcome, while the blockchain layer records the resolved decision and associated metadata. This separation is important because the study examines the effect of supervisory review on decision quality, without treating blockchain execution as the source of detection performance.

\subsection{Role-Aware Configuration Space}

The study evaluates four DS configurations, denoted as DS-1--DS-4. DS-1 assigns centralised ML as the Decider and the LLM as the Supervisor, whereas DS-2 reverses these roles. DS-3 assigns the FedAvg-trained federated meta-model as the Decider and the LLM as the Supervisor, whereas DS-4 reverses these roles. Each configuration is evaluated with both Base and QLoRA LLM variants.

In this paper, \textit{role reversal} denotes a directional comparison in which the same component-family pair is evaluated once in each DS direction. It does not denote a causal intervention in which only the role labels are exchanged while all other mechanisms remain identical. Each direction retains its implemented role-specific candidacy contract, evidence conditions, component outputs, and baseline-derived thresholds. The role-reversal results characterise the measured behaviour of the complete configurations and should not be interpreted as an isolated causal effect of role assignment.

Table~\ref{tab:architecture_definitions} gives the compact architecture map used throughout the paper. The supervisory lifecycle and blockchain audit role are common to all configurations. Primary blocks are retained; only eligible primary approvals become supervisory candidates, unavailable or unreliable evidence preserves the primary approval, and reliable \textsc{Review} or \textsc{Block} outcomes become final blocks under the conservative binary resolver. The resulting lifecycle and final action are recorded with role and execution metadata.

\begin{table}[!t]
\centering
\caption{Role-aware DS configuration map. The arrow denotes Decider $\rightarrow$ Supervisor. C-ML denotes centralised machine learning, and FedAvg denotes the FedAvg-trained federated meta-model.}
\label{tab:architecture_definitions}
\scriptsize
\setlength{\tabcolsep}{2.5pt}
\renewcommand{\arraystretch}{1.12}
\begin{tabularx}{\columnwidth}{@{}L{0.60cm} L{2.35cm} L{1.55cm} Y@{}}
\toprule
\textbf{Cfg.} &
\textbf{Decider $\rightarrow$ Supervisor} &
\textbf{LLM Var.} &
\textbf{Role Test} \\
\midrule
DS-1 &
C-ML $\rightarrow$ LLM &
Base / QLoRA &
Tests LLM supervision over the C-ML primary decision pipeline. \\
\midrule
DS-2 &
LLM $\rightarrow$ C-ML &
Base / QLoRA &
Tests C-ML supervision over the LLM primary decision pipeline. \\
\midrule
DS-3 &
FedAvg $\rightarrow$ LLM &
Base / QLoRA &
Tests LLM supervision over the federated meta-model primary decision pipeline. \\
\midrule
DS-4 &
LLM $\rightarrow$ FedAvg &
Base / QLoRA &
Tests federated meta-model supervision over the LLM primary decision pipeline. \\
\bottomrule
\end{tabularx}
\end{table}

Each configuration is evaluated with the Base and QLoRA variants, with the selected variant applied to the role occupied by the LLM. Algorithm~\ref{alg:role_aware_resolution} shows the decision process used in all four configurations.

\begin{algorithm}[!t]
\footnotesize
\caption{Role-Aware Supervision with Evidence Gating and Conservative Binary Resolution}
\label{alg:role_aware_resolution}
\begin{algorithmic}[1]

\Require Request $x_i$, metadata $\mathbf{m}_i$, identifier $\rho_i$, gate $H$, Decider $D$, Supervisor $S$
\Require Eligibility $G$, validity $R$, thresholds $(\tau_{\mathrm{R}}^{\mathrm{S}},\tau_{\mathrm{B}}^{\mathrm{S}})$, configuration $\kappa$, variant $\nu$, ledger $\mathcal{B}$
\Ensure Final action $a_i^{\mathrm{F}}$ and audit record $b_i$

\State $(\gamma_i,e_i,r_i,u_i)\leftarrow(0,0,0,0)$
\State $p_i^{\mathrm{S}}\leftarrow\varnothing,\quad z_i^{\mathrm{S}}\leftarrow\mathrm{NI}$
\State $h_i\leftarrow H(x_i)$

\If{$h_i=1$}
    \State $(a_i^{\mathrm{P}},a_i^{\mathrm{F}},q_i)\leftarrow
    (\mathrm{BLOCK},\mathrm{BLOCK},\mathrm{HG})$
\Else
    \State $a_i^{\mathrm{P}}\leftarrow D(x_i)$
    \If{$a_i^{\mathrm{P}}=\mathrm{BLOCK}$}
        \State $(a_i^{\mathrm{F}},q_i)\leftarrow(\mathrm{BLOCK},\mathrm{PB})$
    \Else
        \State $\gamma_i\leftarrow G(x_i,\mathbf{m}_i,\kappa,\nu)$
        \If{$\gamma_i=0$}
            \State $(a_i^{\mathrm{F}},q_i)\leftarrow(\mathrm{APPROVE},\mathrm{NS})$
        \Else
            \State $(p_i^{\mathrm{S}},e_i)\leftarrow S(x_i,\mathbf{m}_i)$
            \State $r_i\leftarrow R(p_i^{\mathrm{S}},\mathbf{m}_i,\kappa,\nu)$
            \State $u_i\leftarrow\gamma_i e_i r_i$
            \If{$u_i=0$}
                \State $(z_i^{\mathrm{S}},a_i^{\mathrm{F}},q_i)\leftarrow
                (\mathrm{RETAIN},\mathrm{APPROVE},\mathrm{UR})$
            \ElsIf{$p_i^{\mathrm{S}}\geq\tau_{\mathrm{B}}^{\mathrm{S}}$}
                \State $(z_i^{\mathrm{S}},a_i^{\mathrm{F}},q_i)\leftarrow
                (\mathrm{BLOCK},\mathrm{BLOCK},\mathrm{BE})$
            \ElsIf{$p_i^{\mathrm{S}}\geq\tau_{\mathrm{R}}^{\mathrm{S}}$}
                \State $(z_i^{\mathrm{S}},a_i^{\mathrm{F}},q_i)\leftarrow
                (\mathrm{REVIEW},\mathrm{BLOCK},\mathrm{RE})$
            \Else
                \State $(z_i^{\mathrm{S}},a_i^{\mathrm{F}},q_i)\leftarrow
                (\mathrm{RETAIN},\mathrm{APPROVE},\mathrm{AR})$
            \EndIf
        \EndIf
    \EndIf
\EndIf

\State $\iota_i^{\mathrm{R}}\leftarrow\mathbb{1}\{z_i^{\mathrm{S}}=\mathrm{REVIEW}\}$
\State $\iota_i^{\mathrm{B}}\leftarrow\mathbb{1}\{z_i^{\mathrm{S}}=\mathrm{BLOCK}\}$
\State $\iota_i\leftarrow\iota_i^{\mathrm{R}}+\iota_i^{\mathrm{B}}$
\State $\phi_i\leftarrow\mathrm{ResolverReason}(q_i,\kappa)$
\State $b_i\leftarrow
\{\rho_i,\kappa,\nu,h_i,a_i^{\mathrm{P}},\gamma_i,e_i,u_i,
p_i^{\mathrm{S}},z_i^{\mathrm{S}},\iota_i^{\mathrm{R}},\iota_i^{\mathrm{B}},\iota_i,a_i^{\mathrm{F}},\phi_i\}$
\State $\mathcal{B}.\mathrm{record}(b_i)$
\State \Return $a_i^{\mathrm{F}},b_i$

\end{algorithmic}
\end{algorithm}

In Algorithm~\ref{alg:role_aware_resolution}, $H$ denotes the shared hard-fraud gate, $G$ denotes the role-specific eligibility function, and $R$ denotes the configuration-specific policy-stage evidence-validity function. The parameter $\kappa$ identifies the DS configuration, $\nu$ identifies the LLM variant, and $\rho_i$ denotes the request identifier. The aligned metadata vector $\mathbf{m}_i$ contains the role-specific eligibility and evidence fields required by the selected configuration. The compact lifecycle codes are defined as follows: $\mathrm{HG}$ denotes a hard-gate block, $\mathrm{PB}$ a primary block, $\mathrm{NS}$ an approval without supervision, $\mathrm{UR}$ unavailable or unreliable supervisory evidence, $\mathrm{BE}$ a direct block escalation, $\mathrm{RE}$ a review escalation, and $\mathrm{AR}$ an approval retained after reliable supervision. These codes are used only to keep the pseudocode compact. The configuration-specific resolver reason is denoted by $\phi_i$ and retained in the exported lifecycle artefacts.

The exported artefacts retain the corresponding candidacy, evidence-availability, Supervisor-activation, escalation, intervention, recommendation, and final-action states for each request. This preserves row-level traceability between the formal variables defined above and the implementation outputs. Requests outside the supervisory pathway are distinguished according to whether supervision was not invoked, the hard-fraud gate applied, or the primary decision was already \textsc{Block}.

Algorithm~\ref{alg:role_aware_resolution} applies the decision rules defined in Section~\ref{sec:system_model} and retains review-mediated and direct-block escalations as separate lifecycle events. Because both produce a final block in the current binary pathway, the results report their combined decision effect through the intervention and error-burden measures defined in Section~\ref{sec:experimental_design}. The Supervisor thresholds and their baseline sources are reported in Table~\ref{tab:inherited_thresholds}.

For one request, let $C_{\mathrm{H}}$, $C_{\mathrm{D}}$, $C_{\mathrm{G}}$, $C_{\mathrm{S}}$, $C_{\mathrm{R}}$, and $C_{\mathcal{B}}$ denote the costs of hard-gate evaluation, Decider inference, eligibility evaluation, Supervisor inference, evidence validation, and blockchain recording, respectively. The worst-case time complexity of Algorithm~\ref{alg:role_aware_resolution} is
\[
O\!\left(C_{\mathrm{H}}+C_{\mathrm{D}}+C_{\mathrm{G}}+C_{\mathrm{S}}+C_{\mathrm{R}}+C_{\mathcal{B}}\right),
\]
because each component is invoked at most once per request and the remaining assignments, threshold comparisons, and resolver operations take constant time. For $N$ independently processed requests, the total worst-case complexity is
\[
O\!\left(N\left(C_{\mathrm{H}}+C_{\mathrm{D}}+C_{\mathrm{G}}+C_{\mathrm{S}}+C_{\mathrm{R}}+C_{\mathcal{B}}\right)\right).
\]
Requests resolved by the hard gate or blocked by the Decider avoid Supervisor inference and have lower realised cost. In the deployment replay evaluated in this study, component outputs are precomputed and aligned before policy resolution; consequently, the off-chain eligibility, validation, thresholding, and resolution operations are $O(1)$ per request and $O(N)$ over the replay. With fixed-size lifecycle records, the algorithm requires $O(1)$ working memory per request and $O(N)$ storage for the complete audit output. The model-specific inference costs and blockchain execution overhead are retained explicitly rather than treated as constants; the measured gas, latency, and throughput are reported in Section~\ref{sec:results}.

\subsection{Blockchain Audit and Enforcement Layer}

Once the final action has been resolved, the framework submits the decision record, lifecycle metadata, and retained data and model digests to the blockchain layer. The contract records the request state and governs the permitted review and finalisation transitions without altering the fraud decision itself. The integrity, lifecycle, authorisation, and execution tests are described in Section~\ref{sec:experimental_design}.

\section{Experimental Design}
\label{sec:experimental_design}

This section describes the data, reused component outputs, baseline-derived thresholds, evaluation metrics, operational-validation analyses, and dependability and blockchain tests used in the study.

\subsection{Dataset and Request Labels}

The present study reuses the dataset, preprocessing pipeline, standalone M1--M3 risk-source artefacts, blockchain-linked audit components, and centralised-ML, FedAvg-trained federated meta-model, and LLM-family outputs from the companion study \cite{changizi2026blockchainlinked}; no detector is retrained. The standalone risk sources are denoted as M1 for centralised ML, M2 for the FedAvg-trained federated meta-model, and M3 for the LLM family. Throughout this paper, FedAvg refers to the FedAvg-trained federated meta-model reused from the companion pipeline. Each request is represented by a feature vector and associated metadata. Ground-truth labels distinguish fraudulent requests from legitimate requests, and the operational outcome is binary: \textsc{Approve} or \textsc{Block}.

The reused data are synthetic and were generated by the project-specific data-generation pipeline provided with the companion-study reproducibility materials rather than collected from live network traffic. The training corpus comprises 27 controlled scenario instantiations spanning total anomaly rates of 10\%, 15\%, and 20\% and soft-fraud shares from 20\% to 100\%, producing 27,000 records in total. The independent deployment corpus was generated separately as 100,000 requests from 1,000 synthetic users using a fixed generation seed of 777 and profile seed of 2024, with a 5\% anomaly rate, 10\% of anomalous requests assigned to soft fraud, and a moderate controlled-drift setting. The data-generation and preprocessing code, including the recorded generation parameters, is available through the companion GitLab repository identified in the Data and Code Availability statement.

The raw deployment artefacts contain 100,000 request rows. For label-based predictive and supervisory-intervention evaluation, deployment outputs are aligned with the ground-truth file using the composite business key formed by the hashed identifier, asset hash, timestamp, and Ethereum address. This one-to-one alignment yields 99,996 unique labelled deployment requests after four duplicate-key rows are removed. The blockchain execution analysis instead retains the complete 100,000-row submission stream because it characterises transaction execution and audit overhead rather than label-based predictive performance. Consequently, small differences between the final action rates reported in the predictive tables and the blockchain summary arise from their different denominators and do not indicate inconsistent decision logic.

Within the matched labelled deployment set, a soft-fraud request is defined as a fraudulent request that remains inside the admissible synthetic operating envelope and therefore does not activate the deterministic hard-fraud gate. Formally, soft fraud corresponds to $y_i=1$ and $h_i=0$, whereas hard fraud corresponds to $y_i=1$ and $h_i=1$. The labelled deployment evaluation contains 4,997 fraudulent requests, of which 4,497 are hard-fraud cases resolved by the deterministic gate and 500 are soft-fraud cases that remain available to the AI decision pipeline. Thus, 89.994\% of all fraudulent requests are resolved by the hard gate before Decider--Supervisor intervention. The 500 soft-fraud requests comprise 166 single-component, 167 double-component, and 167 triple-component cases. Soft-fraud recall therefore measures detection performance on the fraudulent requests for which the evaluated AI decision components can affect the outcome.

\subsection{Role Assignment and Score-Transfer Audit}

Role assignments follow Table~\ref{tab:architecture_definitions} and remain unchanged throughout each experiment; no dynamic routing or model selection is used. A matched score-transfer audit checks that the directional comparisons reuse the component evidence from the companion pipeline without recomputing scores for the alternative role direction. Requests are aligned using the canonical composite business key, and each comparison is restricted to requests for which the same component score is available in both implemented directions. Request-level scores are then compared using an absolute numerical tolerance of $1\times10^{-12}$. Ground-truth labels are not used to define the common support or assess score identity. The audit verifies score preservation on matched support, but not equal score availability, equivalent calibration, identical operating thresholds, or a causal effect of role assignment.

\subsection{Baseline-Derived Operating Thresholds}

The main role-aware experiments retain the operating thresholds from the corresponding standalone M1--M3 baseline artefacts. Primary Decider actions are unchanged, while Supervisor review and block thresholds are loaded from the baseline validation-policy artefacts and kept fixed for the main deployment replay. Labelled deployment outcomes are used only for performance assessment and do not determine these reported operating thresholds. In addition, a separate conditional-calibration sensitivity analysis re-estimates an escalation boundary using only the held-out validation requests that the Decider approves, satisfy the configuration-specific candidacy contract, have available Supervisor evidence, and contain a finite Supervisor score. This supplementary analysis is used to test whether the main findings can be explained solely by transferring thresholds from the standalone population to the conditional supervisory population; it does not replace the predefined operating policy.

Table~\ref{tab:inherited_thresholds} reports the Supervisor thresholds used in the four role-aware configurations. The Base and QLoRA LLM Supervisors use variant-specific thresholds from the M3 LLM baseline. The centralised-ML and FedAvg Supervisors use the M1 and M2 thresholds, respectively. Because the centralised-ML and FedAvg Supervisors are unchanged across the two LLM-primary variants, their thresholds remain identical for Base and QLoRA.

\begin{table}[!t]
\centering
\caption{Baseline-derived Supervisor thresholds used by the role-aware configurations.}
\label{tab:inherited_thresholds}

\scriptsize
\renewcommand{\arraystretch}{1.16}
\setlength{\tabcolsep}{3.2pt}

\resizebox{\columnwidth}{!}{%
\begin{tabular}{@{}c c c c l@{}}
\toprule

\textbf{Cfg.} &
\textbf{Supervisor} &
\shortstack{\textbf{LLM}\\\textbf{Variant}} &
\shortstack{\textbf{Thresholds}\\
$\boldsymbol{\tau_{\mathrm{R}}^{\mathrm{S}}/\tau_{\mathrm{B}}^{\mathrm{S}}}$} &
\textbf{Threshold Source} \\

\midrule

DS-1 &
LLM &
Base &
$0.166912/0.213826$ &
M3 Base-LLM baseline \\

DS-1 &
LLM &
QLoRA &
$0.108244/0.290672$ &
M3 QLoRA-LLM baseline \\

\midrule

DS-2 &
C-ML &
Base/QLoRA &
$0.111644/0.397767$ &
M1 C-ML baseline \\

\midrule

DS-3 &
LLM &
Base &
$0.166912/0.213826$ &
M3 Base-LLM baseline \\

DS-3 &
LLM &
QLoRA &
$0.108244/0.290672$ &
M3 QLoRA-LLM baseline \\

\midrule

DS-4 &
Federated meta-model &
Base/QLoRA &
$0.096649/0.272116$ &
M2 FedAvg-trained baseline \\

\bottomrule
\end{tabular}%
}

\vspace{1mm}
\parbox{\columnwidth}{%
\scriptsize
\textit{Note:} Thresholds are reported as review/block. C-ML denotes centralised machine learning, LLM denotes large language model, QLoRA denotes Quantized Low-Rank Adaptation, and FedAvg denotes the FedAvg-trained federated meta-model. All values in this table are taken from the corresponding standalone baseline validation-policy artefacts and constitute the predefined main-policy thresholds. A separate validation-only conditional-calibration sensitivity analysis is reported in Section~\ref{subsec:conditional_calibration}.
}

\end{table}

The component scores are reused without modification from the corresponding companion pipelines, where they are defined as probability-valued fraud-risk outputs. Their probability provenance and unit-interval support are verified for the present analysis. However, equivalent calibration across the centralised-ML, FedAvg, Base-LLM, and QLoRA score families is not assumed. Calibration measures such as the Brier score, expected calibration error, reliability curves, calibration slope, and calibration intercept are not re-estimated in this operational study. The score-based analyses concern within-component threshold, ranking, and sensitivity; they do not support cross-family probability-calibration comparisons.

\subsection{Conditional Validation Calibration Sensitivity}
\label{subsec:conditional_calibration}

To assess whether threshold transfer from the standalone baselines explains the supervisory results, an additional validation-only calibration analysis is performed on the conditional supervisory population. For each configuration and LLM variant, calibration uses only validation requests for which the primary action is \textsc{Approve}, the configuration-specific candidate flag is active, Supervisor evidence is available, and the Supervisor score is finite. Candidate escalation thresholds are drawn from the observed conditional-validation scores, together with a no-intervention sentinel above the maximum observed score.

Three prespecified validation rules are evaluated independently: a minimum-error rule that minimises total validation errors, a maximum-F1 rule that maximises validation F1, and a false-positive-budget rule requiring no more than 50 added false positives per recovered fraud case when such a point exists. These rules provide a favourable conditional operating test for supervision by selecting the escalation boundary on the exact validation subpopulation eligible for supervisory intervention rather than transferring the standalone-population threshold unchanged. Threshold selection uses validation data only. Each selected threshold is then frozen and replayed on the labelled deployment population without further optimisation. Because the implemented binary resolver maps both \textsc{Review} and \textsc{Block} escalation to a final block, the binary outcome identifies an escalation boundary but does not independently identify two deployment-optimal review and block thresholds. The inherited block threshold is therefore retained for lifecycle interpretation, while the conditional analysis is treated as a sensitivity analysis rather than a replacement operating policy.

\subsection{LLM Variants and Role Use}

Both Base and QLoRA variants are evaluated in every configuration. In DS-1 and DS-3, the LLM is the Supervisor, so the primary decision is unchanged across variants and any difference appears only after supervision. In DS-2 and DS-4, the LLM is the Decider; Base and QLoRA produce different primary profiles, while the same variant has the same primary output before the centralised-ML or FedAvg Supervisor is applied. These comparisons characterise the evaluated pipelines and do not attribute the results solely to QLoRA adaptation.

\subsection{Decision Stages}

Each configuration is evaluated before supervision using the primary decision $a_i^{\mathrm{P}}$ and after resolution using the final decision $a_i^{\mathrm{F}}$, as defined in Section~\ref{sec:system_model}. Their difference quantifies the change associated with supervision in the matched deployment replay. Review-mediated and direct-block escalations remain separate in the lifecycle records, although both lead to a final \textsc{Block} under the implemented binary resolver.

\subsection{Evaluation Metrics}

The evaluation reports accuracy, precision, recall, and F1 score using their standard binary-classification definitions, together with FPR, block rate, total errors, and soft-fraud recall. True positives (TP), false positives (FP), true negatives (TN), and false negatives (FN) follow their standard binary-classification meanings. The FPR is defined as the fraction of legitimate requests incorrectly blocked:
\[
\mathrm{FPR} = \frac{\mathrm{FP}}{\mathrm{FP}+\mathrm{TN}}.
\]

The block rate measures the fraction of all requests blocked by the decision policy:
\[
\mathrm{Block\ Rate} = \frac{\mathrm{TP}+\mathrm{FP}}{N},
\]
where $N$ is the total number of requests. Total errors are computed as the sum of false positives and false negatives:
\[
\mathrm{Total\ Errors} = \mathrm{FP}+\mathrm{FN}.
\]

Let $\mathcal{S}=\{i:y_i=1,\ h_i=0\}$ denote the labelled soft-fraud subset. Soft recall at decision stage $d\in\{\mathrm{P},\mathrm{F}\}$ is defined as

\begin{equation}
\mathrm{Soft\ Recall}^{(d)}
=
\frac{
\sum_{i=1}^{N}
\mathbb{1}
\left\{
i\in\mathcal{S}
\land
a_i^{(d)}=\mathrm{BLOCK}
\right\}
}{
|\mathcal{S}|
},
\label{eq:soft_recall}
\end{equation}

where $a_i^{(\mathrm{P})}$ and $a_i^{(\mathrm{F})}$ denote the primary-only and final supervised actions, respectively. In the matched labelled deployment evaluation, $|\mathcal{S}|=500$. Soft recall is reported separately because high aggregate fraud recall may conceal missed in-range fraud cases that are not captured by the deterministic hard gate.

The primary performance and intervention tables report deterministic point estimates on the matched deployment replay of 99,996 labelled requests. Request-level sampling uncertainty is assessed with 10,000-replicate paired stratified bootstrap analyses that preserve pairing and the legitimate, soft-fraud, and hard-fraud strata. The primary bootstrap estimates within-configuration changes from the primary-only to supervised resolved decisions. A secondary paired bootstrap evaluates the implemented role-direction contrasts for the centralised-ML--LLM and FedAvg--LLM pairs on the same matched requests. The resulting intervals are treated as effect-estimation intervals rather than multiplicity-adjusted discovery tests. Traffic-mix, Review-capacity, conditional-threshold, and fault/attack analyses are treated as exploratory sensitivity or stress-test families; no pointwise significance claims are made across their multiple scenarios. Accordingly, the reported comparisons characterise the measured replay and its request-sampling uncertainty without establishing population-level superiority across retraining runs, datasets, or deployment environments.

\subsection{Supervisor Intervention Metrics}

The row-level intervention indicator $\iota_i$ is defined in Section~\ref{sec:system_model}. Its aggregate rate is

\begin{equation}
\mathrm{Intervention\ Rate}
=
\frac{\sum_{i=1}^{N}\iota_i}{N}.
\label{eq:intervention_rate}
\end{equation}

Review-mediated and direct-block escalations remain separate in the lifecycle records, but the main intervention tables report their sum because both convert a primary approval into a final block.

True-positive gain counts fraudulent requests blocked only after supervision, while false positives added counts legitimate requests newly blocked by supervision:

\[
\mathrm{TP\ Gain} =
\#\{i:y_i=1,\ a_i^{\mathrm{P}}=\mathrm{APPROVE},\ a_i^{\mathrm{F}}=\mathrm{BLOCK}\},
\]

\[
\mathrm{FP\ Added} =
\#\{i:y_i=0,\ a_i^{\mathrm{P}}=\mathrm{APPROVE},\ a_i^{\mathrm{F}}=\mathrm{BLOCK}\}.
\]

Their ratio is

\[
\mathrm{FP\ per\ TP\ Gain} =
\frac{\mathrm{FP\ Added}}{\mathrm{TP\ Gain}}.
\]

The ratio is reported as not applicable when $\mathrm{TP\ Gain}=0$. The analysis also reports $\Delta\mathrm{Recall}$, $\Delta\mathrm{Soft\ Recall}$, $\Delta\mathrm{FPR}$, $\Delta\mathrm{Block\ Rate}$, $\Delta\mathrm{F1}$, and $\Delta\mathrm{Total\ Errors}$ between the primary and final stages. These measures characterise the measured fraud-recovery and legitimate-user-blocking trade-off; they do not represent a monetary or operator-specific utility function.

Because the shared hard-fraud gate blocks hard-fraud requests before supervision, every supervisory true-positive gain occurs within the 500-request soft-fraud subset. The corresponding soft-recall change is

\begin{equation}
\Delta\mathrm{Soft\ Recall}\;(\mathrm{pp})
=
100
\frac{\mathrm{TP\ Gain}}{500}.
\label{eq:soft_recall_change}
\end{equation}

This identity is also used to check consistency between the intervention counts and the reported soft-recall changes.

\subsection{Blockchain Evaluation Setup}

Blockchain execution is evaluated over the complete 100,000-row submission stream described above. The recorded lifecycle and contract controls follow the framework defined in Section~\ref{sec:framework}. The exported measures include final approval and block rates, average gas per transaction, mean and 95th-percentile end-to-end latency, measured throughput, validator participation rate, and Jain fairness index. The main results table reports the principal outcome-rate and execution-overhead measures, while the participation and fairness outputs remain available in the reproducibility artefacts.

\subsection{Reproducibility Artefacts}

The experimental pipeline exports the row-level decisions, lifecycle fields, comparison summaries, robustness outputs, blockchain test records, figures, and LaTeX tables used in the paper. Each artefact is indexed by decision stage, configuration, LLM variant, and experiment, allowing reported results to be traced to the corresponding source output. Tables and figures are generated from the exported files rather than assembled manually.

The retained validation outputs also include the population-alignment and score-semantics checks, traffic-mix and \textsc{Review}-capacity results, score-transfer audit, Pareto and cost-sensitivity analyses, validation manifests, and hashed output inventory. Repository access is described in the Data and Code Availability statement.

\subsection{Operational Traffic-Mix and Review-Capacity Validation}
\label{subsec:operational_validation}

Operational sensitivity is evaluated over the matched population of 99,996 requests using the retained primary and final actions, candidacy states, Supervisor scores, thresholds, and resolver outputs. These request-level values remain unchanged throughout the analysis; only the assumed traffic composition or available \textsc{Review} capacity is varied.

Traffic composition is varied through exact class-conditional reweighting. For each configuration and variant, block rates are estimated separately for legitimate, soft-fraud, and hard-fraud requests. Fraud prevalence takes values of 0.5\%, 1\%, 2\%, 5\%, 10\%, and 20\%, while the soft-fraud share of the fraudulent population takes values of 0\%, 10\%, 25\%, 50\%, 75\%, and 100\%. The resulting 36 scenarios preserve the measured class-conditional decision behaviour and report primary and final outcomes as rates and expected events per 100,000 requests. Labels are used to estimate these class-conditional outcomes, but not to alter thresholds or request-level decisions.

Finite \textsc{Review} capacity is assessed by reconstructing the Review-band population from the retained Supervisor scores, lifecycle flags, and review and block thresholds. Candidates are ranked by decreasing Supervisor score, with the deterministic row identifier used to break ties. Labels are excluded from queue construction and ranking and are introduced only after selection to measure fraud yield, legitimate-request burden, severity composition, queue coverage, and captured queued fraud. The evaluated capacities are 0, 10, 25, 50, 100, 250, 500, 1,000, 1,500, 2,000, and 2,500 requests per 100,000 deployment requests.

The three-way analysis keeps \textsc{Approve}, \textsc{Block}, and unresolved \textsc{Review} separate and evaluates fail-open and fail-closed treatments of queue overflow. Perfect adjudication is reported only as an oracle upper bound for selected Review cases; it does not represent an implemented reviewer or measured human performance.

The 16 primary and final operating points are also compared using Pareto dominance over false-negative and false-positive counts, with duplicate objective pairs collapsed before frontier construction. A supplementary resource-aware comparison includes selected Review volume. Normalised cost sensitivity uses

\begin{equation}
C_j=
\lambda_{\mathrm{FN}}\mathrm{FN}_j
+\mathrm{FP}_j
+\lambda_{\mathrm{R}} R_j,
\label{eq:normalised_cost}
\end{equation}

where the false-positive weight is fixed to one, $\lambda_{\mathrm{FN}}$ is the assumed relative cost of a missed fraud case, $\lambda_{\mathrm{R}}$ is the assumed relative cost of selecting a Review case, and $R_j$ is the selected Review volume. These ratios are analytical assumptions rather than estimated financial costs, and the corresponding break-even regions are reported separately from the oracle Review bounds.

\subsection{Dependability, Robustness, and Security Assessment}

All tests use the matched deployment replay and the frozen DS-1--DS-4 lifecycle artefacts; no model is retrained, no LLM inference is rerun, and no policy is retuned on the labelled deployment set. Request-sampling uncertainty is estimated using 10,000 paired stratified bootstrap replicates with 95\% confidence intervals. Primary and final decisions remain paired, and resampling preserves the legitimate, soft-fraud, and hard-fraud stratum sizes. These intervals capture sampling variation within the fixed replay, not uncertainty from retraining, alternative random seeds, regenerated data, or new deployment environments.

Policy sensitivity is examined by jointly shifting the baseline-derived Supervisor review and block thresholds from $-0.10$ to $+0.10$ around their reported values. The conservative resolver is also compared with a direct-block-only alternative, while a separate three-way interpretation leaves \textsc{Review} unresolved to measure queue volume and composition. Component faults cover Supervisor-evidence unavailability, LLM parsing and fallback failures, bounded Supervisor-score perturbations, and hypothetical primary-LLM failure-resolution strategies. Experiments that sample intermediate fault locations use 500 replicates per fault rate.

Targeted score manipulation is evaluated under the K2-O oracle class-targeted profile, which assumes exact knowledge of the review threshold and target class. Bounded upward and downward shifts are applied only to eligible threshold-crossing cases while primary decisions, candidacy conditions, thresholds, and resolver logic remain unchanged. Labels are used only to construct class-specific worst-case evaluation sets, not as information available during ordinary operation. The test measures oracle-targeted threshold manipulation and does not establish general adversarial robustness.

Blockchain security and lifecycle integrity are tested through digest reconstruction, controlled post-export payload modification, duplicate and conflicting initial-decision submissions, repeated review resolution and finalisation, post-finalisation state mutation, and unauthorised calls to \texttt{recordAIDecision} and \texttt{resolveReview}. Live tests use fresh deployments of the retained contract application binary interface (ABI) and bytecode. Duplicate submissions are sent as new Ethereum transactions with new account nonces, so they test application-level duplicate handling; replay of an identical signed transaction is not evaluated. Authorisation tests compare unauthorised calls with authorised-validator controls and inspect both transaction outcomes and request-state changes.

Table~\ref{tab:fault_threat_mapping} links the formal threat and fault model to the controlled assessments. The expected-property column states the behaviour under test without presuming that the property was satisfied. Test outcomes and violations are reported in Section~\ref{sec:results}.

\begin{table*}[!t]
\centering
\caption{Mapping of the evaluated faults and threats to their test scenarios and evaluation criteria. K2-O denotes the oracle class-targeted profile with exact knowledge of the review threshold and target class.}
\label{tab:fault_threat_mapping}

\footnotesize
\renewcommand{\arraystretch}{1.16}
\setlength{\tabcolsep}{4pt}

\begin{tabularx}{\textwidth}{@{}L{1.35cm} L{2.9cm} L{4.8cm} Y@{}}
\toprule
\textbf{Category} &
\textbf{Test Target} &
\textbf{Test Scenario} &
\textbf{Evaluation Criterion} \\
\midrule

\multirow{6}{1.35cm}{\centering\textbf{Faults}}
& Supervisor evidence
& Evidence unavailable at controlled fault rates
& Assess whether unavailable evidence preserves the primary decision and existing primary blocks without creating additional blocks. \\

& Supervisor score
& Systematic additive score bias within the evaluated perturbation range
& Quantify changes in supervisory intervention, fraud recovery, and legitimate-user blocking under bounded score corruption. \\

& LLM output interface
& Parsing failure or fallback activation
& Assess whether invalid LLM output is prevented from being treated as reliable evidence or producing an unrecorded escalation. \\

& Primary LLM
& Primary output unavailable under alternative resolution strategies
& Compare the consequences of explicit failure-resolution strategies without treating hypothetical alternatives as validated production failover mechanisms. \\

\midrule

\multirow{16}{1.35cm}{\centering\textbf{Threats}}
& Exported audit payload
& Post-export modification of retained decision fields
& Assess whether payload modification is detectable through digest recomputation. \\

& Initial-decision endpoint
& Exact duplicate initial-decision submission
& Assess duplicate handling, including whether repeated submission changes request state or re-emits decision events. \\

& Initial-decision endpoint
& Conflicting pre-finalisation decision submission
& Assess whether a conflicting resubmission can alter the stored request state before finalisation. \\

& Finalisation endpoint
& Duplicate finalisation attempt
& Assess whether repeated finalisation is rejected and the finalised request state is preserved. \\

& Review-resolution endpoint
& Repeated or conflicting review-resolution submission
& Assess whether review resolution is single-use and subsequent resolution attempts are rejected. \\

& Finalised decision state
& Attempted decision mutation after finalisation
& Assess whether the finalised decision state is protected against subsequent modification. \\

& Finalised review state
& Attempted review mutation after finalisation
& Assess whether the finalised review state can be reopened or modified. \\

& Supervisor score and thresholds
& K2-O targeted threshold-crossing manipulation under bounded budgets
& Quantify the operational effect of class-targeted threshold crossings under oracle knowledge without implying threshold-agnostic robustness. \\

& Protected contract endpoints
& Unauthorised calls to \texttt{recordAIDecision} and \texttt{resolveReview}
& Assess whether unauthorised calls are rejected without request-state change while the corresponding authorised controls remain executable. \\

\bottomrule
\end{tabularx}
\end{table*}

\section{Results and Analysis}
\label{sec:results}

Results are reported for DS-1--DS-4 across predictive performance, supervisory intervention, operational sensitivity, dependability, and blockchain execution. The main comparison is between each Decider's primary output and the corresponding final decision after supervision.

\subsection{Hard-Gate Dominance and Non-Hard Decision Performance}
\label{subsec:nonhard_performance}

Because the deterministic hard gate resolves the large majority of fraudulent requests before Decider--Supervisor intervention, the aggregate fraud-recall values over the complete matched population are strongly influenced by deterministic hard-gate decisions. To isolate the requests on which the evaluated AI components can affect outcomes, Table~\ref{tab:nonhard_performance} reports performance on the 95,499 non-hard requests, comprising 94,999 legitimate requests and 500 soft-fraud requests.

\begin{table*}[!t]
\centering
\caption{Primary-only and supervised resolved performance on the 95,499 non-hard deployment requests. Percentage-based metrics are reported in \%. The population contains 94,999 legitimate requests and 500 soft-fraud requests; the 4,497 hard-fraud requests resolved by the deterministic gate are excluded.}
\label{tab:nonhard_performance}
\scriptsize
\renewcommand{\arraystretch}{1.12}
\setlength{\tabcolsep}{3.0pt}

\begin{tabular*}{\textwidth}{@{\extracolsep{\fill}}lllrrrrr@{}}
\toprule
\textbf{Config.} &
\textbf{Variant} &
\textbf{Stage} &
\textbf{Precision} &
\textbf{Recall} &
\textbf{F1} &
\textbf{FPR} &
\textbf{Total Errors} \\
\midrule

\multirow{4}{*}{DS-1}
& \multirow{2}{*}{Base}
& Primary-only & 2.58 & 83.00 & 5.01 & 16.46 & 15726 \\
&
& Supervised   & 1.01 & 91.80 & 2.01 & 47.18 & 44864 \\
& \multirow{2}{*}{QLoRA}
& Primary-only & 2.58 & 83.00 & 5.01 & 16.46 & 15726 \\
&
& Supervised   & 2.58 & 83.20 & 5.01 & 16.53 & 15783 \\

\midrule

\multirow{4}{*}{DS-2}
& \multirow{2}{*}{Base}
& Primary-only & 1.01 & 75.60 & 1.99 & 39.15 & 37314 \\
&
& Supervised   & 1.09 & 91.00 & 2.16 & 43.38 & 41256 \\
& \multirow{2}{*}{QLoRA}
& Primary-only & 2.35 & 82.40 & 4.57 & 18.01 & 17202 \\
&
& Supervised   & 1.88 & 86.60 & 3.67 & 23.84 & 22717 \\

\midrule

\multirow{4}{*}{DS-3}
& \multirow{2}{*}{Base}
& Primary-only & 8.67 & 70.00 & 15.43 & 3.88 & 3838 \\
&
& Supervised   & 1.04 & 88.80 & 2.06 & 44.45 & 42286 \\
& \multirow{2}{*}{QLoRA}
& Primary-only & 8.67 & 70.00 & 15.43 & 3.88 & 3838 \\
&
& Supervised   & 8.15 & 70.80 & 14.62 & 4.20 & 4134 \\

\midrule

\multirow{4}{*}{DS-4}
& \multirow{2}{*}{Base}
& Primary-only & 1.01 & 75.60 & 1.99 & 39.15 & 37314 \\
&
& Supervised   & 1.11 & 88.80 & 2.19 & 41.60 & 39575 \\
& \multirow{2}{*}{QLoRA}
& Primary-only & 2.35 & 82.40 & 4.57 & 18.01 & 17202 \\
&
& Supervised   & 2.09 & 85.20 & 4.09 & 20.98 & 20000 \\

\bottomrule
\end{tabular*}
\end{table*}

On the non-hard population, the contrast between additional fraud recovery and legitimate-request burden becomes substantially clearer. DS-3 QLoRA remains the least disruptive supervised profile: soft-fraud recall rises from 70.0\% to 70.8\%, while F1 falls from 15.43\% to 14.62\% and total errors increase from 3,838 to 4,134. By contrast, DS-1 Base and DS-3 Base substantially increase soft-fraud recall but do so through large increases in false positives, reducing non-hard F1 to 2.01\% and 2.06\%, respectively. The non-hard analysis is therefore treated as the primary assessment of the AI decision layer; complete-population results are retained below to show the end-to-end system outcome including deterministic hard-gate decisions.

\subsection{Full Matched-Population Context}

For end-to-end context, Table~\ref{tab:primary_supervised} reports primary and final performance on the complete matched labelled replay, including the 4,497 hard-fraud requests resolved by the deterministic gate. The primary results are identical across Base and QLoRA in DS-1 and DS-3 because their Deciders are unchanged. Within each LLM variant, DS-2 and DS-4 likewise share the same LLM-primary result before their respective Supervisors are applied.

\begin{table*}[!t]
\centering
\caption{Primary-only versus supervised resolved decision performance on 99,996 unique labelled deployment requests. Percentage-based metrics are reported in \%.}
\label{tab:primary_supervised}

\scriptsize
\renewcommand{\arraystretch}{1.15}
\setlength{\tabcolsep}{2pt}

\begin{tabular*}{\textwidth}{@{\extracolsep{\fill}}lllrrrrrrrr@{}}
\toprule
\textbf{Config.} &
\textbf{Variant} &
\textbf{Stage} &
\textbf{Acc.} &
\textbf{Prec.} &
\textbf{Recall} &
\textbf{F1} &
\textbf{Soft Recall} &
\textbf{FPR} &
\textbf{Block Rate} &
\textbf{Total Errors} \\
\midrule

\multirow{4}{*}{DS-1}
& \multirow{2}{*}{Base}
& Primary-only & 84.27 & 23.90 & 98.30 & 38.45 & 83.00 & 16.46 & 20.55 & 15726 \\
&
& Supervised   & 55.13 & 9.96  & 99.18 & 18.10 & 91.80 & 47.18 & 49.78 & 44864 \\

& \multirow{2}{*}{QLoRA}
& Primary-only & 84.27 & 23.90 & 98.30 & 38.45 & 83.00 & 16.46 & 20.55 & 15726 \\
&
& Supervised   & 84.22 & 23.84 & 98.32 & 38.37 & 83.20 & 16.53 & 20.61 & 15783 \\

\midrule

\multirow{4}{*}{DS-2}
& \multirow{2}{*}{Base}
& Primary-only & 62.68 & 11.59 & 97.56 & 20.72 & 75.60 & 39.15 & 42.07 & 37314 \\
&
& Supervised   & 58.74 & 10.73 & 99.10 & 19.36 & 91.00 & 43.38 & 46.16 & 41256 \\

& \multirow{2}{*}{QLoRA}
& Primary-only & 82.80 & 22.29 & 98.24 & 36.34 & 82.40 & 18.01 & 22.02 & 17202 \\
&
& Supervised   & 77.28 & 17.88 & 98.66 & 30.27 & 86.60 & 23.84 & 27.58 & 22717 \\

\midrule

\multirow{4}{*}{DS-3}
& \multirow{2}{*}{Base}
& Primary-only & 96.16 & 56.79 & 97.00 & 71.64 & 70.00 & 3.88  & 8.54  & 3838 \\
&
& Supervised   & 57.71 & 10.47 & 98.88 & 18.94 & 88.80 & 44.45 & 47.17 & 42286 \\

& \multirow{2}{*}{QLoRA}
& Primary-only & 96.16 & 56.79 & 97.00 & 71.64 & 70.00 & 3.88  & 8.54  & 3838 \\
&
& Supervised   & 95.87 & 54.88 & 97.08 & 70.12 & 70.80 & 4.20  & 8.84  & 4134 \\

\midrule

\multirow{4}{*}{DS-4}
& \multirow{2}{*}{Base}
& Primary-only & 62.68 & 11.59 & 97.56 & 20.72 & 75.60 & 39.15 & 42.07 & 37314 \\
&
& Supervised   & 60.42 & 11.11 & 98.88 & 19.98 & 88.80 & 41.60 & 44.46 & 39575 \\

& \multirow{2}{*}{QLoRA}
& Primary-only & 82.80 & 22.29 & 98.24 & 36.34 & 82.40 & 18.01 & 22.02 & 17202 \\
&
& Supervised   & 80.00 & 19.81 & 98.52 & 32.99 & 85.20 & 20.97 & 24.85 & 20000 \\

\bottomrule
\end{tabular*}
\end{table*}

Among the supervised outputs, DS-3 QLoRA has the highest F1 and the lowest FPR, at 70.12\% and 4.20\%, respectively. Relative to its own FedAvg primary stage, however, it gains only 0.08 percentage points in recall and 0.80 percentage points in soft recall, while F1 falls by 1.52 percentage points and total errors increase by 296. DS-3 Base is substantially more disruptive: F1 falls from 71.64\% to 18.94\%, and FPR rises from 3.88\% to 44.45\%.

DS-2 produces decision-changing interventions in both variants. The centralised-ML Supervisor changes 4,096 Base decisions and 5,557 QLoRA decisions, increasing recall by 1.54 and 0.42 percentage points, respectively. These gains are accompanied by higher FPRs, lower F1 scores, and increases of 3,942 and 5,515 total errors relative to the corresponding LLM-primary outputs. Across all eight configuration--variant pipelines, supervision increases recall but lowers F1 and raises total errors relative to the corresponding primary stage.

\begin{figure*}[!t]
    \centering
    \includegraphics[width=0.92\textwidth]{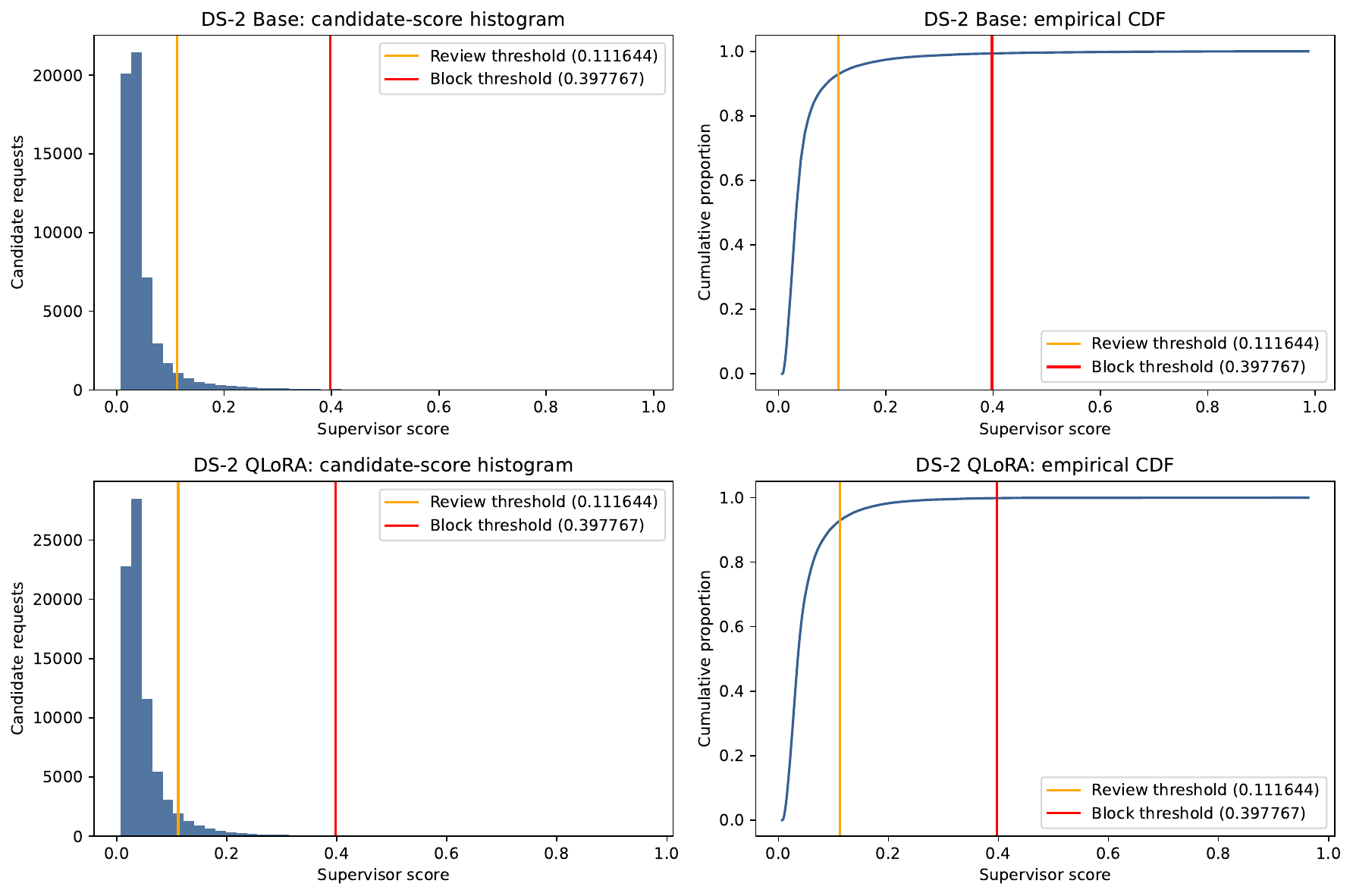}
    \caption{DS-2 centralised-ML Supervisor score diagnostics for the Base and QLoRA LLM-primary variants. Histograms and empirical cumulative-distribution functions show the conditional candidate-score distributions. Vertical lines mark the inherited review threshold (0.111644) and block threshold (0.397767). Both candidate populations contain fully non-degenerate scores, and both contain observations crossing the inherited intervention thresholds.}
    \label{fig:ds2_score_diagnostics}
\end{figure*}

Fig.~\ref{fig:ds2_score_diagnostics} reports the additional check of the DS-2 result against the centralised-ML Supervisor score distribution, motivated by an earlier pipeline version that produced no decision-changing DS-2 interventions. The corrected candidate populations contain 57,929 finite Base scores and 77,973 finite QLoRA scores, with every observed score unique within each population; the distributions are therefore not degenerate. Under the inherited centralised-ML review and block thresholds of 0.111644 and 0.397767, respectively, 4,096 Base candidates and 5,557 QLoRA candidates lie at or above the review threshold, including 379 Base and 97 QLoRA candidates at or above the block threshold. The resulting DS-2 interventions therefore reflect threshold crossings in a non-degenerate score distribution rather than a constant or collapsed Supervisor output.

\subsection{Supervisor Intervention and Error Trade-Offs}

Table~\ref{tab:supervisor_impact} reports the decision-changing interventions and the associated true-positive gains, false positives added, and changes in recall, soft recall, and FPR for each configuration and variant.

\begin{table*}[!t]
\centering
\caption{Supervisor intervention and error trade-offs over 99,996 unique labelled deployment requests. Intervention rate is reported in \%, while metric changes are reported in percentage points (pp).}
\label{tab:supervisor_impact}

\scriptsize
\renewcommand{\arraystretch}{1.15}
\setlength{\tabcolsep}{2pt}

\begin{tabular*}{\textwidth}{@{\extracolsep{\fill}}llrrrrrrrr@{}}
\toprule
\textbf{Config.} &
\textbf{Variant} &
\textbf{Interv. Count} &
\textbf{Interv. Rate} &
\textbf{TP Gain} &
\textbf{FP Added} &
\textbf{FP per TP Gain} &
\textbf{$\Delta$Recall} &
\textbf{$\Delta$Soft Recall} &
\textbf{$\Delta$FPR} \\
\midrule

\multirow{2}{*}{DS-1}
& Base  & 29226 & 29.23 & 44 & 29182 & 663.23 & 0.88 & 8.80 & 30.72 \\
& QLoRA & 59    & 0.06  & 1  & 58    & 58.00  & 0.02 & 0.20 & 0.06 \\

\midrule

\multirow{2}{*}{DS-2}
& Base  & 4096 & 4.10 & 77 & 4019 & 52.19  & 1.54 & 15.40 & 4.23 \\
& QLoRA & 5557 & 5.56 & 21 & 5536 & 263.62 & 0.42 & 4.20  & 5.83 \\

\midrule

\multirow{2}{*}{DS-3}
& Base  & 38636 & 38.64 & 94 & 38542 & 410.02 & 1.88 & 18.80 & 40.57 \\
& QLoRA & 304   & 0.30  & 4  & 300   & 75.00  & 0.08 & 0.80  & 0.32 \\

\midrule

\multirow{2}{*}{DS-4}
& Base  & 2393 & 2.39 & 66 & 2327 & 35.26  & 1.32 & 13.20 & 2.45 \\
& QLoRA & 2826 & 2.83 & 14 & 2812 & 200.86 & 0.28 & 2.80  & 2.96 \\

\bottomrule
\end{tabular*}
\end{table*}

Because every supervisory true-positive gain occurs among the 500 soft-fraud requests, exact two-sided Clopper--Pearson intervals were calculated for the corresponding soft-fraud recovery proportions. The observed $\Delta$Soft Recall values and 95\% intervals are 8.80 [6.47, 11.63] percentage points for DS-1 Base, 0.20 [0.01, 1.11] for DS-1 QLoRA, 15.40 [12.35, 18.87] for DS-2 Base, 4.20 [2.62, 6.35] for DS-2 QLoRA, 18.80 [15.47, 22.51] for DS-3 Base, 0.80 [0.22, 2.04] for DS-3 QLoRA, 13.20 [10.36, 16.49] for DS-4 Base, and 2.80 [1.54, 4.65] for DS-4 QLoRA. Ratios based on fewer than five recovered fraud cases are especially unstable: the FP-per-TP-gain values of 58.00 for DS-1 QLoRA and 75.00 for DS-3 QLoRA are based on only one and four additional true positives, respectively, and should not be interpreted as stable operating ratios.

The largest intervention burdens occur in DS-1 Base and DS-3 Base. They add 29,182 and 38,542 false positives for 44 and 94 additional true positives, respectively. Using QLoRA as the LLM Supervisor sharply reduces intervention volume in DS-1 and DS-3, but still produces substantially more false positives than true-positive gains, with FP-per-TP-gain ratios of 58.00 and 75.00. In DS-2, the centralised-ML Supervisor adds 4,019 false positives for 77 additional true positives in Base and 5,536 false positives for 21 additional true positives in QLoRA. DS-4 Base has the lowest FP-per-TP-gain ratio, at 35.26, although its 66 additional true positives are accompanied by 2,327 additional false positives.

\begin{figure}[!t]
    \centering
    \includegraphics[width=\linewidth]{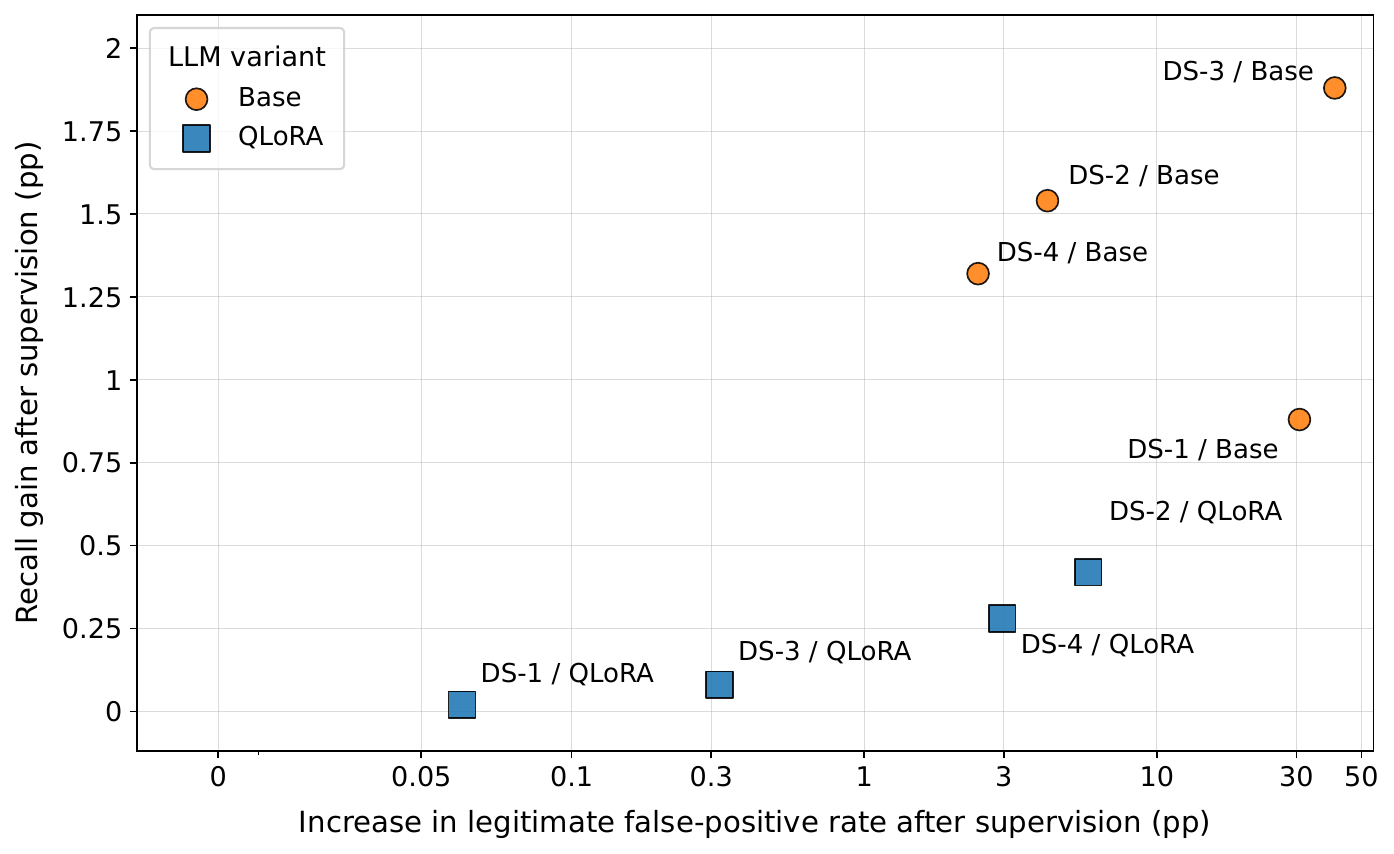}
    \caption{$\Delta$Recall versus $\Delta$FPR across role-aware DS configurations.}
    \label{fig:delta_tradeoff}
\end{figure}

Fig.~\ref{fig:delta_tradeoff} shows that all eight supervised profiles increase both recall and FPR. For DS-2, the centralised-ML Supervisor increases recall by 1.54 percentage points for Base and 0.42 percentage points for QLoRA, while increasing FPR by 4.23 and 5.83 percentage points, respectively. The figure separates the scale of the recall gain from the accompanying legitimate-request burden; a higher intervention rate is not treated as a benefit.

\subsection{Directional Role Comparison}

Table~\ref{tab:role_reversal} compares the two implemented role directions for each model-family pair and LLM variant using the supervised resolved outputs and paired-bootstrap intervals. The better-direction label identifies the configuration with the stronger measured value for the reported metric, while the gap and corresponding 95\% confidence interval quantify the request-level directional difference. These comparisons concern complete implemented directional configurations and are not interpreted as isolated causal effects of exchanging the two roles.

\begin{table}[!t]
\centering
\caption{Paired-bootstrap role-direction gap summary for the predefined DS configurations. Gaps and 95\% confidence intervals are reported in percentage points (pp). Each gap is expressed as the magnitude favouring the configuration shown in the Better Config. column. C-ML denotes centralised ML.}
\label{tab:role_reversal}

\scriptsize
\renewcommand{\arraystretch}{1.15}
\setlength{\tabcolsep}{1.8pt}

\begin{tabular*}{\columnwidth}{@{\extracolsep{\fill}}l c l c c@{}}
\toprule
\textbf{Pair} &
\textbf{Var.} &
\textbf{Metric} &
\textbf{Better Config.} &
\textbf{Gap [95\% CI]} \\
\midrule

C-ML $\leftrightarrow$ LLM &
Base &
F1 &
DS-2 &
1.26 [1.22, 1.31] \\

C-ML $\leftrightarrow$ LLM &
Base &
FPR &
DS-2 &
3.80 [3.68, 3.92] \\

\midrule

C-ML $\leftrightarrow$ LLM &
QLoRA &
F1 &
DS-1 &
8.10 [7.89, 8.33] \\

C-ML $\leftrightarrow$ LLM &
QLoRA &
FPR &
DS-1 &
7.32 [7.14, 7.49] \\

\midrule

FedAvg $\leftrightarrow$ LLM &
Base &
F1 &
DS-4 &
1.04 [0.98, 1.10] \\

FedAvg $\leftrightarrow$ LLM &
Base &
FPR &
DS-4 &
2.85 [2.71, 3.00] \\

\midrule

FedAvg $\leftrightarrow$ LLM &
QLoRA &
F1 &
DS-3 &
37.13 [36.53, 37.72] \\

FedAvg $\leftrightarrow$ LLM &
QLoRA &
FPR &
DS-3 &
16.78 [16.55, 17.01] \\

\bottomrule
\end{tabular*}
\end{table}

The paired-bootstrap intervals support substantial directional separation for most reported F1 and FPR contrasts within the matched replay. For the centralised-ML--LLM pair, DS-2 has the stronger Base profile: its F1 advantage is 1.26 percentage points (95\% CI 1.22--1.31), and its FPR is lower by 3.80 percentage points (95\% CI 3.68--3.92). For QLoRA, the direction reverses: DS-1 has an F1 advantage of 8.10 percentage points (95\% CI 7.89--8.33) and an FPR advantage of 7.32 percentage points (95\% CI 7.14--7.49). For the FedAvg--LLM pair, DS-4 has the stronger Base F1 and FPR profile, whereas the largest separation occurs for QLoRA: DS-3 exceeds DS-4 in F1 by 37.13 percentage points (95\% CI 36.53--37.72) and has an FPR lower by 16.78 percentage points (95\% CI 16.55--17.01). These are secondary request-level effect-estimation intervals for the complete implemented directional configurations; they do not isolate a causal effect of exchanging role labels or account for retraining and deployment uncertainty.

The score-transfer audit found no request-level mismatch across the eight comparisons at the $1\times10^{-12}$ tolerance. The matched support contained 95,499 requests for each centralised-ML--LLM pair, 57,929 for the Base FedAvg--LLM pair, and 77,973 for the QLoRA FedAvg--LLM pair. Hard-fraud requests were absent because they were resolved before both component scores became available. The result verifies score identity only on matched support; role-specific candidacy, score availability, thresholds, and resolver contracts remain different.

\subsection{Decision Transition and Block-Rate Shift}

\begin{figure}[!t]
    \centering
    \includegraphics[width=\linewidth]{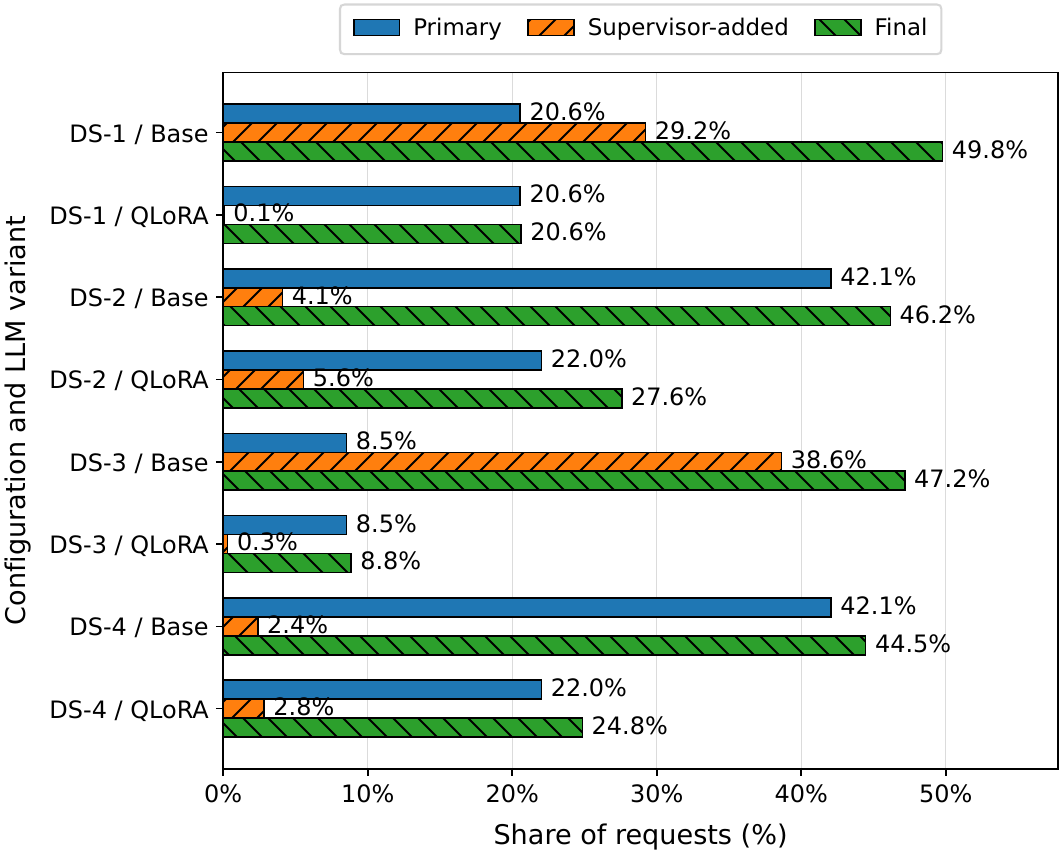}
    \caption{Primary block rate, Supervisor-added block rate, and final block rate across role-aware configurations.}
    \label{fig:transition}
\end{figure}

Fig.~\ref{fig:transition} decomposes each final block rate into the primary contribution and the additional blocks introduced by supervision. The largest block-rate shifts occur in DS-1 Base and DS-3 Base, where supervision adds 29.23 and 38.64 percentage points, respectively. The corresponding QLoRA shifts are substantially smaller, at 0.06 percentage points for DS-1 and 0.30 percentage points for DS-3. DS-2 shows block-rate shifts of 4.10 percentage points for Base and 5.56 percentage points for QLoRA. DS-4 adds 2.39 and 2.83 percentage points for Base and QLoRA, respectively.

\subsection{Three-Way Review Interpretation and Review Workload}
\label{subsec:three_way_review}

\begin{figure}[!t]
    \centering
    \includegraphics[width=\columnwidth]{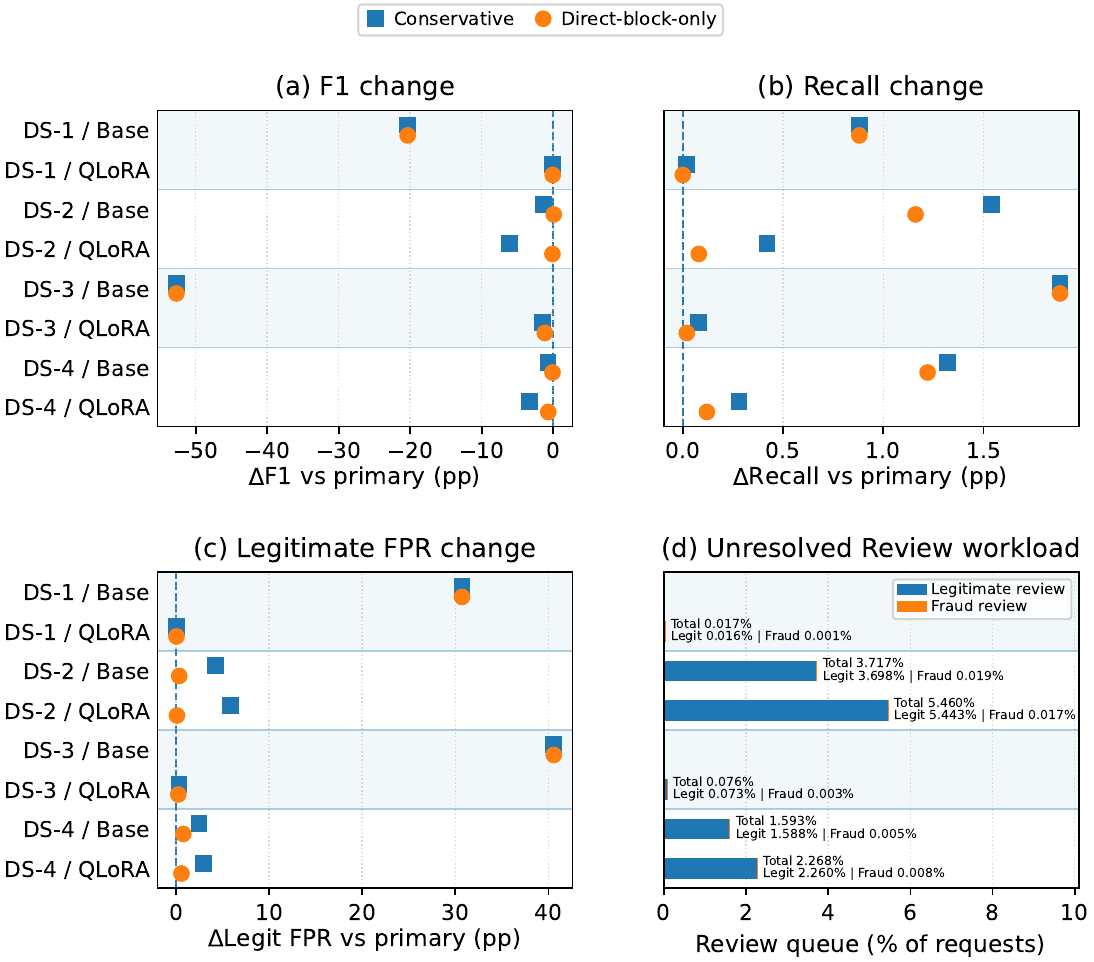}
    \caption{Comparison of the conservative and direct-block-only resolvers. Panels (a)--(c) report changes in F1 score, fraud recall, and FPR relative to the primary-only stage. Panel (d) reports the unresolved \textsc{Review} workload under a three-way interpretation of \textsc{Review}, decomposed into legitimate and fraudulent requests as percentages of all deployment requests.}
    \label{fig:alternative_resolver}
\end{figure}

Fig.~\ref{fig:alternative_resolver} compares the implemented conservative resolver with a direct-block-only alternative and a three-way interpretation in which \textsc{Review} remains unresolved. Under the conservative resolver, both \textsc{Review} and \textsc{Block} become final blocks; under the direct-block-only alternative, only \textsc{Block} changes a primary approval. The three-way interpretation instead routes \textsc{Review} cases to an unresolved queue. These are analytical policy comparisons; the implemented blockchain pathway remains binary.

The direct-block-only resolver reduces legitimate false-positive growth and improves F1 in the six pipelines containing non-zero \textsc{Review} escalations. Nevertheless, all eight intervention-producing pipelines remain below their corresponding primary-stage F1 scores. The disadvantage cannot be attributed solely to mapping \textsc{Review} to a final block, because direct Supervisor \textsc{Block} outcomes also contribute to the error profile.

Non-zero \textsc{Review} queues occur in six pipelines. DS-1 QLoRA contains 17 cases, including one fraudulent request, and DS-3 QLoRA contains 76 cases, including three fraudulent requests. DS-2 produces substantially larger queues: 3,717 cases for Base, comprising 19 fraudulent and 3,698 legitimate requests, and 5,460 cases for QLoRA, comprising 17 fraudulent and 5,443 legitimate requests. The DS-4 queues contain 1,593 cases for Base and 2,268 for QLoRA, but only five and eight fraudulent requests, respectively. Thus, the DS-2 and DS-4 queues are overwhelmingly dominated by legitimate requests. Treating \textsc{Review} as unresolved can reduce immediate blocking while creating a low-yield operational queue. These results describe queue size and composition only; reviewer accuracy, delay, cost, and post-review outcomes are not evaluated.

\subsection{Traffic-Mix and Finite-Capacity Review Sensitivity}
\label{subsec:traffic_review_sensitivity}

The class-conditional reweighting reproduced the canonical deployment baseline for all eight configuration--variant artefacts. Table~\ref{tab:traffic_mix_sensitivity} summarises how the supervised final output compares with the corresponding primary-only stage across the 36 traffic-mix scenarios.

\begin{table}[!t]
\caption{Primary-to-final outcomes across the 36 traffic-mix scenarios. Counts indicate the number of scenarios in which the supervised final output is higher, equal to, or lower than the corresponding primary-only output. For total errors, lower values indicate improvement.}
\label{tab:traffic_mix_sensitivity}
\centering
\scriptsize
\setlength{\tabcolsep}{4.2pt}
\renewcommand{\arraystretch}{1.12}
\begin{tabular}{llccc@{\hspace{10pt}}ccc}
\toprule
\textbf{Cfg.} &
\textbf{LLM Variant} &
\multicolumn{3}{c}{\textbf{F1 Score vs. Primary}} &
\multicolumn{3}{c}{\textbf{Total Errors vs. Primary}} \\
\cmidrule(lr){3-5}
\cmidrule(lr){6-8}
&
&
\textbf{Higher} &
\textbf{Equal} &
\textbf{Lower} &
\textbf{Lower} &
\textbf{Equal} &
\textbf{Higher} \\
\midrule
DS-1 & Base  & 0  & 0  & 36 & 0 & 0  & 36 \\
DS-1 & QLoRA & 3  & 0  & 33 & 0 & 0  & 36 \\
DS-2 & Base  & 14 & 0 & 22 & 0 & 0 & 36 \\
DS-2 & QLoRA & 0  & 0 & 36 & 0 & 0 & 36 \\
DS-3 & Base  & 0  & 0  & 36 & 0 & 0  & 36 \\
DS-3 & QLoRA & 1  & 0  & 35 & 0 & 0  & 36 \\
DS-4 & Base  & 18 & 0  & 18 & 2 & 0  & 34 \\
DS-4 & QLoRA & 0  & 0  & 36 & 0 & 0  & 36 \\
\bottomrule
\end{tabular}
\end{table}

Across the 36 traffic mixtures, DS-1 Base, DS-2 QLoRA, DS-3 Base, and DS-4 QLoRA always show lower F1 and more total errors after supervision. DS-1 QLoRA, DS-2 Base, and DS-3 QLoRA achieve higher F1 in three, 14, and one scenarios, respectively, but never reduce total errors. DS-4 Base is the only pipeline to reduce errors: this occurs in two scenarios with 20\% fraud prevalence and soft fraud comprising 75\% or 100\% of the fraudulent population, yielding reductions of approximately 20.4 and 680.4 errors per 100,000 requests. In the other 34 scenarios, including every scenario with fraud prevalence at or below 10\%, supervision increases total errors. DS-4 Base nevertheless achieves higher F1 in 18 scenarios, showing that a higher F1 score does not necessarily correspond to fewer incorrect decisions.

Finite-capacity analysis shows that a capacity of 100 reviews per 100,000 requests is sufficient to clear the smaller DS-1 QLoRA and DS-3 QLoRA queues, yielding only one and three fraud cases, respectively. The larger DS-2 and DS-4 queues remain low-yield under the tested capacities. For DS-2 Base, the top 100 requests capture three of the 19 queued fraud cases, while for DS-2 QLoRA they capture one of 17. Processing 2,500 requests captures 13 fraud cases for DS-2 Base and 11 for DS-2 QLoRA; complete recovery requires processing their full queues of 3,717 and 5,460 requests. For DS-4 Base, the top 100 requests capture three of the five queued fraud cases and the top 500 capture four; the fifth appears only when the complete 1,593-request queue is processed. For DS-4 QLoRA, the top 100 contain no fraud, the top 500 capture four of eight cases, and the top 1,000 capture seven; processing all 2,268 requests is required to recover the eighth.

The full-queue fraud yields are 0.511\% for DS-2 Base, 0.311\% for DS-2 QLoRA, 0.314\% for DS-4 Base, and 0.353\% for DS-4 QLoRA. Score-based prioritisation concentrates some fraud near the front of the queues but does not remove the dominant legitimate-request burden. These counts describe queue composition after label-free ranking; they do not measure reviewer accuracy, delay, cost, or final adjudication. Perfect adjudication remains an oracle upper bound, and \textsc{Review} capacity, overflow policy, and post-review authority would need to be specified before deployment.

\subsection{Pareto and Normalised Cost-Ratio Sensitivity}
\label{subsec:pareto_cost_sensitivity}

After duplicate objective pairs are collapsed, the 16 primary and final candidates form 10 unique false-negative--false-positive operating points, seven of which lie on the Pareto frontier. The primary FedAvg output has 150 false negatives and 3,688 false positives, whereas the DS-3 QLoRA final output has 146 false negatives and 3,988 false positives. This shift prevents four additional fraud misses at the cost of 300 additional legitimate blocks. Adding selected \textsc{Review} volume as a third objective leaves the same 10 unique points and seven frontier points.

With the false-positive cost normalised to one and the \textsc{Review} cost set to zero, the DS-3 QLoRA final output becomes lower-cost than the primary FedAvg output only when $\lambda_{\mathrm{FN}}>75$; the two are equal at $\lambda_{\mathrm{FN}}=75$. This break-even value is derived from the measured replay outputs and is not an empirically estimated operator cost or a deployment-optimal threshold. Oracle \textsc{Review} resolution remains a non-deployable upper bound.

\subsection{Robustness, Fault Tolerance, and Policy Sensitivity}

\begin{figure}[!t]
    \centering
    \includegraphics[width=\columnwidth]{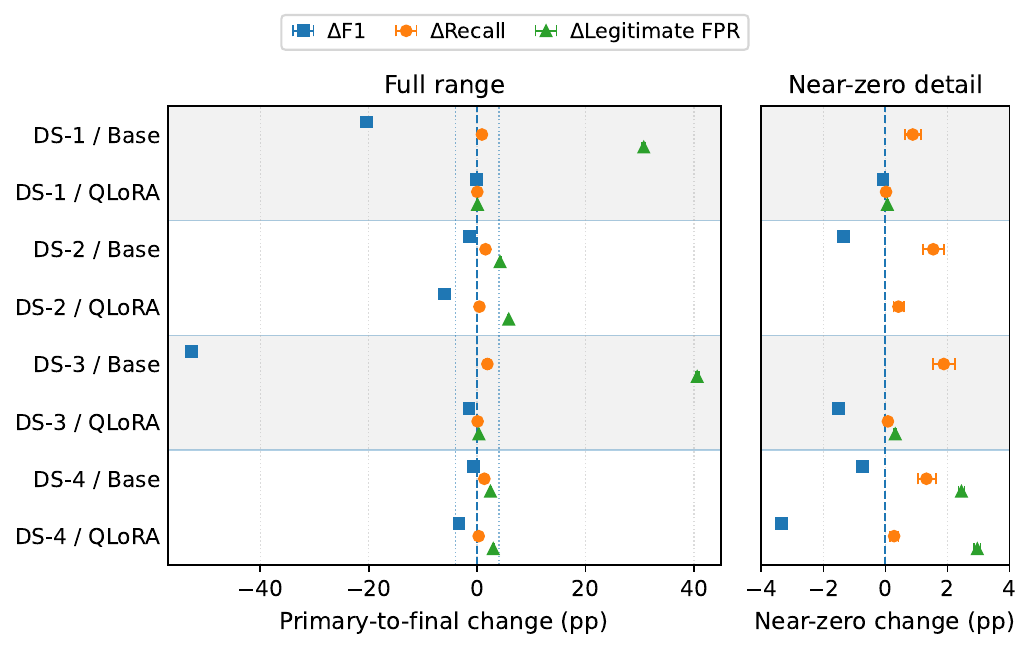}
    \caption{Paired-bootstrap uncertainty intervals for the primary-to-final changes in F1 score, fraud recall, and FPR. Points show the measured changes, and horizontal intervals show 95\% confidence intervals obtained from 10,000 paired stratified bootstrap replicates. The magnified panel highlights changes close to zero.}
    \label{fig:bootstrap_robustness}
\end{figure}

Fig.~\ref{fig:bootstrap_robustness} reports paired-bootstrap intervals for the primary-to-final changes. Across all eight intervention-producing pipelines, the 95\% interval for the F1 change remains below zero and that for the FPR change remains above zero. The small recall increase for DS-1 QLoRA includes zero at the lower confidence limit.

For DS-3 QLoRA, the measured recall increase is 0.08 percentage points, with a 95\% interval of 0.02 to 0.16 percentage points. Its F1 change remains negative, from $-1.70$ to $-1.34$ percentage points, while the FPR increase ranges from 0.28 to 0.35 percentage points. The additional fraud recovery therefore does not offset the deterioration in F1 and legitimate-user preservation. These intervals quantify request-level sampling variation within the matched replay.

\begin{figure}[!t]
    \centering
    \includegraphics[width=\columnwidth]{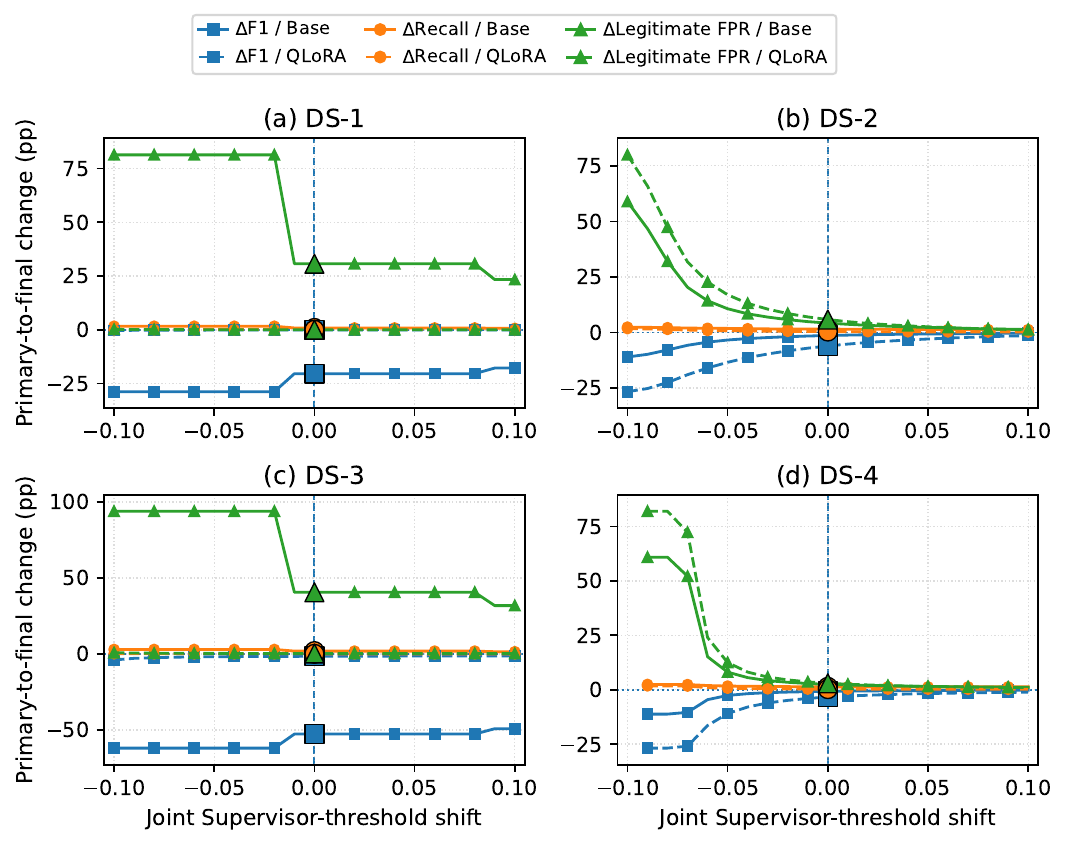}
    \caption{Local joint-threshold sensitivity of the primary-to-final changes in F1 score, fraud recall, and FPR. Supervisor review and block thresholds are shifted jointly from $-0.10$ to $+0.10$ around their baseline operating values. Larger black-edged markers at $x=0$ identify the reported operating points.}
    \label{fig:threshold_sensitivity}
\end{figure}

Fig.~\ref{fig:threshold_sensitivity} reports the local threshold-sensitivity analysis, which jointly shifts the baseline-derived Supervisor review and block thresholds from $-0.10$ to $+0.10$ while keeping component outputs, candidacy conditions, and resolver rules unchanged.

Across all valid tested points, no intervention-producing pipeline matches or exceeds its primary-stage F1 or reduces total errors to the primary-stage level. Lower shifts generally increase intervention, recall, and FPR, whereas higher shifts move the outputs towards the corresponding primary decisions. This pattern applies to DS-2 as well as to the other directional configurations. The disadvantage persists across the tested local range and is not confined to the baseline-derived threshold pair. This is a post hoc sensitivity analysis, not threshold optimisation, and no operating point is selected using the labelled deployment results.

The validation-only conditional-calibration analysis provides a stronger test of threshold dependence than the local threshold shifts because candidate thresholds are re-derived directly from the conditional validation population. The resulting deployment replay does not show a stable improvement over the corresponding primary-only decisions. Under the minimum-error rule, DS-2 Base increases F1 by 0.137 percentage points and recall by 1.121 percentage points, but also increases FPR by 0.182 percentage points and total errors by 117. DS-2 QLoRA instead decreases F1 by 0.114 percentage points and adds 99 errors. The corresponding DS-4 Base and QLoRA minimum-error replays change F1 by +0.019 and -0.300 percentage points and increase total errors by 423 and 243, respectively. Across the validation-selected minimum-error and maximum-F1 operating points, none reduces deployment total errors below its corresponding primary-only stage. The false-positive-budget rule is additionally unstable across configurations and, in several cases, produces near-universal blocking when replayed on deployment data. These results indicate that conditional calibration was examined explicitly but did not yield a consistently transferable supervisory advantage; the baseline-derived thresholds are therefore retained as the predefined main-policy results rather than being retrospectively replaced by a deployment-favourable calibrated setting.

\begin{figure}[!t]
    \centering
    \includegraphics[width=\columnwidth]{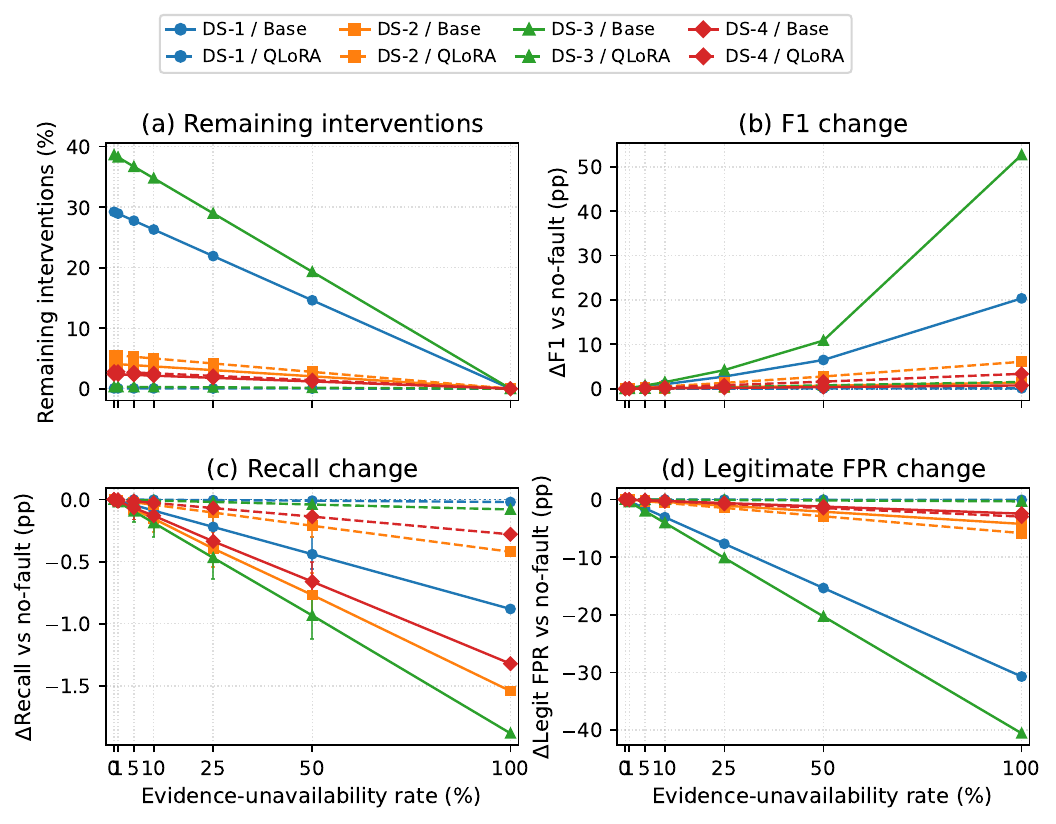}
    \caption{Response of the role-aware pipelines to increasing Supervisor-evidence unavailability. Panel (a) reports the remaining supervisory-intervention rate, while panels (b)--(d) report changes in F1 score, fraud recall, and FPR relative to the no-fault condition. Lines show the median outcomes, with uncertainty intervals shown for fault rates evaluated through repeated fault-location sampling.}
    \label{fig:evidence_unavailability}
\end{figure}

Fig.~\ref{fig:evidence_unavailability} reports the response to Supervisor-evidence unavailability injected at rates of 0\%, 1\%, 5\%, 10\%, 25\%, 50\%, and 100\%, while primary decisions, candidacy conditions, thresholds, and resolver rules remain unchanged. As evidence becomes unavailable, affected requests lose supervisory intervention and retain their primary decisions. At complete unavailability, all eight configuration--variant pipelines reproduce their primary-only binary decisions exactly, eliminate all supervisory interventions, preserve every existing primary block, and create no new blocks. This confirms fail-safe recovery to the Decider output under the tested contract, but not continued supervisory availability or preservation of the additional soft-fraud recovery previously introduced by supervision.

\begin{figure}[!t]
    \centering
    \includegraphics[width=\columnwidth]{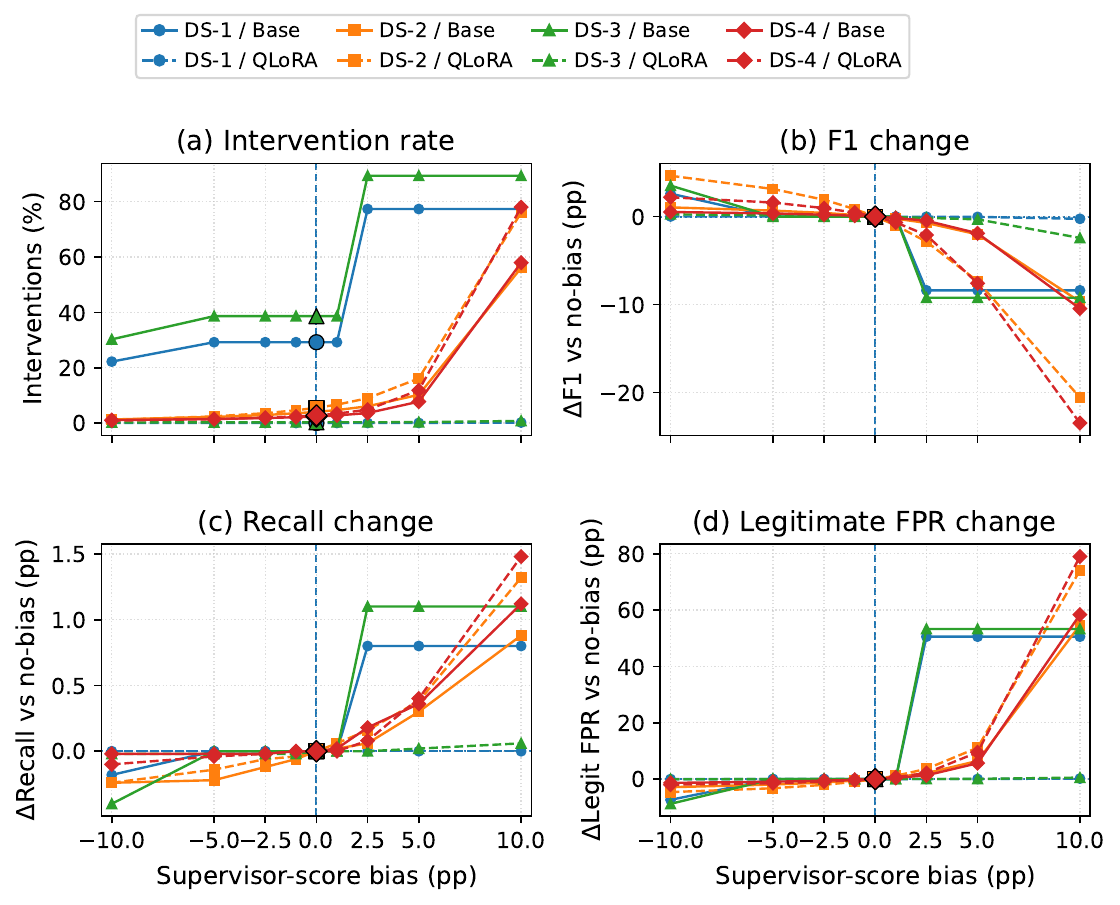}
    \caption{Sensitivity of the role-aware pipelines to additive Supervisor-score bias. Panel (a) reports the supervisory-intervention rate, while panels (b)--(d) report changes in F1 score, fraud recall, and FPR relative to the no-bias operating point. Larger black-edged markers at $x=0$ identify the reported no-bias operating points.}
    \label{fig:score_bias}
\end{figure}

Fig.~\ref{fig:score_bias} reports deterministic additive biases of $-0.10$, $-0.05$, $-0.025$, $-0.01$, $0$, $+0.01$, $+0.025$, $+0.05$, and $+0.10$ applied to the available Supervisor scores and clipped to $[0,1]$. Primary decisions, candidacy conditions, thresholds, and resolver rules remain unchanged.

Higher Supervisor-score biases produce non-decreasing intervention, final-block, true-positive, and false-positive counts. At a bias of $+0.05$, DS-3 Base intervenes on 89,320 requests, raises FPR by 53.29 percentage points, and adds 50,574 total errors relative to the no-bias condition. At $+0.10$, DS-4 QLoRA intervenes on 77,973 requests, raises FPR by 79.03 percentage points, and adds 74,999 total errors.

These results show that supervisory behaviour depends jointly on score distributions, intervention thresholds, and resolver policy. The injected biases are deterministic stress-test conditions; they do not estimate the likelihood or magnitude of score corruption in deployment.

The K2-O experiment applies bounded class-targeted score shifts only to eligible requests whose Supervisor scores can cross the relevant intervention boundary. It assumes exact knowledge of the baseline-derived review threshold and target class, while labels are used only to construct worst-case evaluation sets. The analysis is an oracle upper bound; it does not represent an ordinary attacker capability or establish general adversarial robustness.

At an attack budget of $0.10$, legitimate-targeted upward shifts change 48,041 and 50,629 legitimate decisions in DS-1 Base and DS-3 Base, increasing FPR by 50.57 and 53.29 percentage points, respectively. DS-2 is also exposed: 51,914 legitimate decisions change for Base and 70,394 for QLoRA, increasing FPR by 54.65 and 74.10 percentage points and reducing F1 by 9.75 and 20.69 percentage points, respectively. The largest effect occurs in DS-4 QLoRA, where 75,073 legitimate requests cross the intervention boundary, FPR rises by 79.03 percentage points, and F1 falls by 23.60 percentage points relative to the unattacked baseline. Fraud-targeted downward effects are smaller because fewer fraudulent requests occupy eligible threshold-crossing positions.

\begin{figure*}[!t]
    \centering
    \includegraphics[width=0.95\textwidth]{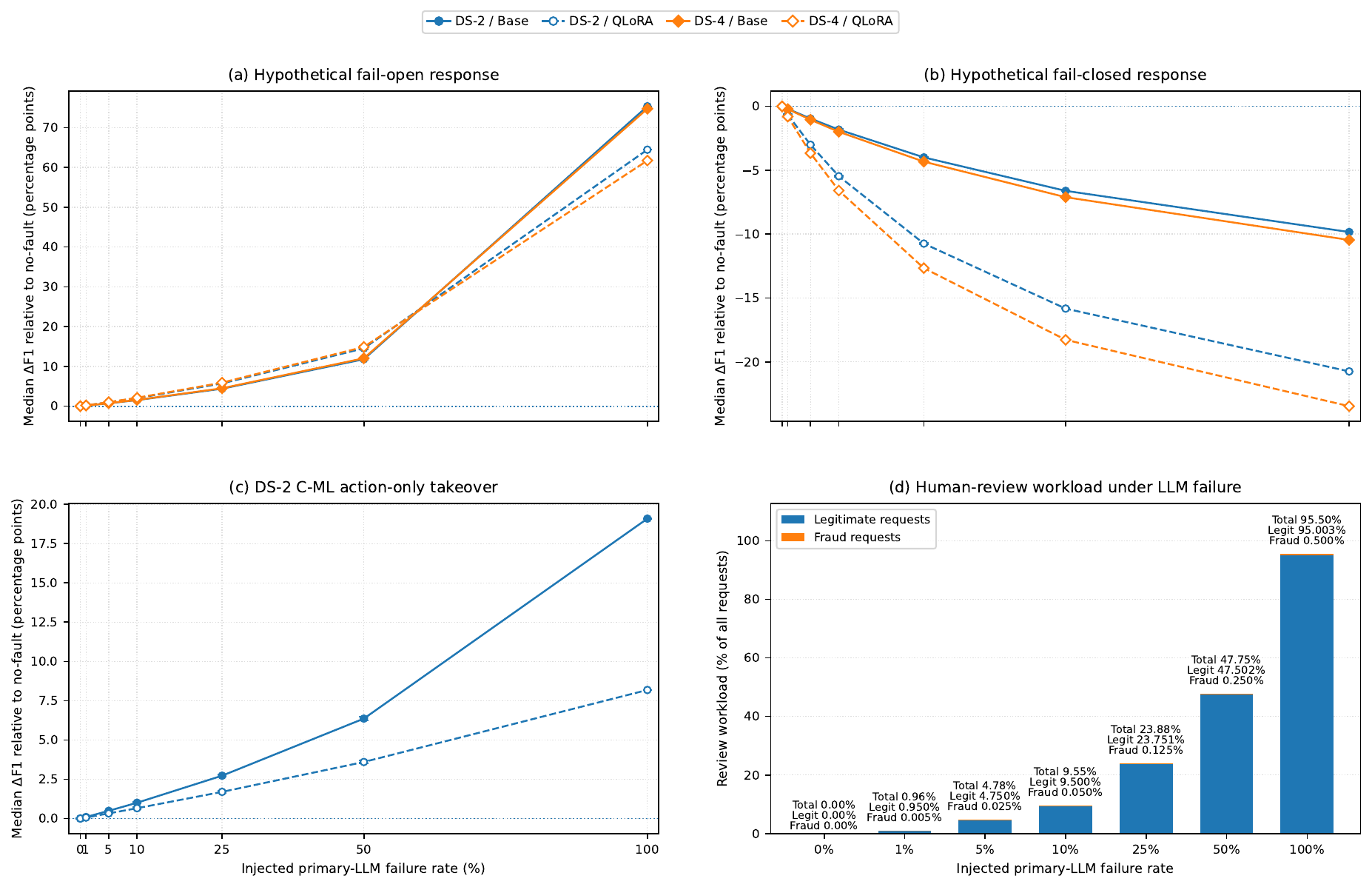}
    \caption{Sensitivity of the LLM-primary pipelines to hypothetical primary-output failure-resolution strategies. Panels (a)--(c) report the median F1-score change under fail-open resolution, fail-closed resolution, and the hypothetical DS-2 centralised-ML action-only takeover, respectively. Error bars denote the central 95\% simulation interval across random fault-location replicates; where these intervals are narrower than the plotted markers, they are not visually distinguishable. Panel (d) reports the human-review workload under LLM failure, decomposed into total, legitimate, and fraudulent requests.}
    \label{fig:primary_llm_failure}
\end{figure*}

The primary-LLM failure analysis examines DS-2 and DS-4, where the LLM produces the primary action. Because the retained lifecycle artefacts do not define a run-time contract for unavailable primary-LLM output, fail-open, fail-closed, unresolved \textsc{Review}, and a DS-2 centralised-ML action-only takeover are evaluated only as hypothetical post hoc strategies. Fig.~\ref{fig:primary_llm_failure} compares their effects.

Fail-open resolution reduces blocking but increases exposure to missed fraud, whereas fail-closed resolution protects against missed fraud at the cost of substantial legitimate-request blocking. Leaving affected requests unresolved avoids fabricating a binary outcome but can create a large \textsc{Review} queue dominated by legitimate requests in the evaluated replay.

For DS-2, the centralised-ML action-only takeover strategy substitutes the stored centralised-ML binary action for the unavailable LLM decision. Because the retained artefacts do not provide an independently estimated centralised-ML confidence measure or an explicit takeover contract, this result does not represent an implemented failover mechanism. An equivalent FedAvg takeover is not evaluated for DS-4 because complete secondary evidence is unavailable for the original LLM-block population.

Table~\ref{tab:robustness_summary} provides a compact summary of the principal uncertainty, policy-sensitivity, and fault-injection results and the scope of each analysis.

\begin{table*}[!t]
\centering
\caption{Summary of robustness, policy-sensitivity, and fault-injection results.}
\label{tab:robustness_summary}
\scriptsize
\setlength{\tabcolsep}{3pt}
\renewcommand{\arraystretch}{1.14}

\begin{tabularx}{\textwidth}{@{}L{2.15cm} L{3.05cm} X X@{}}
\toprule
\textbf{Assessment} &
\textbf{Test} &
\textbf{Key Result} &
\textbf{Scope} \\
\midrule

Paired request uncertainty &
10,000 paired stratified bootstrap replicates with 95\% confidence intervals. &
F1-change intervals were below zero and FPR-change intervals above zero for all eight intervention-producing pipelines. &
Quantifies request-sampling uncertainty within the matched deployment replay, rather than uncertainty from retraining, alternative corpus generation, or deployment conditions. \\

\midrule

Threshold sensitivity &
Joint shifts of the baseline-derived Supervisor review and block thresholds from $-0.10$ to $+0.10$. &
No tested intervention-producing operating point matched or exceeded its primary stage on F1 while also achieving no greater total errors. &
Local post hoc sensitivity analysis; labelled deployment results were not used to select or retune thresholds. \\

\midrule

Resolver sensitivity &
Conservative resolution was compared with direct-block-only resolution and an unresolved three-way \textsc{Review} interpretation. &
Alternative \textsc{Review} handling reduced false-positive burden, but no decision-changing pipeline matched its primary stage on both F1 and total errors. &
Quantifies the effect and workload of alternative \textsc{Review} handling without modelling reviewer decisions or post-review outcomes. \\

\midrule

Supervisor-evidence failure &
Evidence-unavailability and LLM parse/fallback faults from 0\% to 100\%, with repeated fault-location sampling at intermediate rates. &
Complete evidence loss exactly recovered the primary-only decisions, protected existing primary blocks, and created no new blocks. &
Tests fail-safe retention of the primary action when supervisory evidence is unavailable or invalid. \\

\midrule

Supervisor-score perturbation &
Additive score biases from $-0.10$ to $+0.10$, with perturbed Supervisor scores clipped to $[0,1]$. &
Increasing positive bias increased interventions, false positives, and total errors and could sharply increase legitimate-user blocking. &
Bounded deterministic stress test of score corruption rather than an estimate of its operational probability. \\

\midrule

Targeted threshold-crossing manipulation &
K2-O oracle class-targeted upward and downward score shifts under bounded attack budgets. &
Upward manipulation could sharply increase legitimate-user blocking; at budget $0.10$, DS-4 QLoRA produced 75,073 additional false positives and a 79.03-percentage-point FPR increase. &
Uses ground-truth labels only to construct worst-case target populations and does not represent deployment-time label access. \\

\midrule

Primary-LLM failure &
Hypothetical fail-open, fail-closed, unresolved-\textsc{Review}, and DS-2 centralised-ML action-only takeover strategies. &
The resulting operating profile depended strongly on the assumed failure-resolution strategy. &
Post hoc failure-policy sensitivity rather than implemented run-time failover behaviour. \\

\bottomrule
\end{tabularx}
\end{table*}

\subsection{Blockchain Auditability and Enforcement Summary}

Table~\ref{tab:blockchain_summary} summarises the blockchain-recorded final actions and execution overhead for the complete 100,000-transaction stream in each profile. C-ML denotes centralised ML; final approval/block rates and end-to-end latency are reported as approval/block and average/P95, respectively.

\begin{table}[!t]
\centering
\caption{Blockchain auditability and enforcement summary over the complete 100,000 submitted transactions per execution profile. Final approval/block rates are reported in \%, end-to-end latency as average and 95th-percentile (P95) values in milliseconds, and throughput in transactions per second.}
\label{tab:blockchain_summary}

\scriptsize
\renewcommand{\arraystretch}{1.20}
\setlength{\tabcolsep}{1.5pt}

\begin{tabular*}{\columnwidth}{@{\extracolsep{\fill}}llrrrr@{}}
\toprule
\textbf{Cfg.} &
\textbf{Variant} &
\shortstack{\textbf{Final}\\\textbf{Appr./Block}} &
\shortstack{\textbf{Avg Gas/}\\\textbf{Transaction}} &
\shortstack{\textbf{Latency}\\\textbf{Avg./P95}} &
\textbf{Throughput} \\
\midrule

\multirow{2}{*}{DS-1}
& Base  & 50.217/49.783 & 1154200 & 114.2903/146.2 & 7.4556 \\
& QLoRA & 79.384/20.616 & 1162669 & 144.3108/202.2 & 5.9492 \\

\midrule

\multirow{2}{*}{DS-2}
& Base  & 53.833/46.167 & 1174526 & 127.0726/170.8 & 6.8344 \\
& QLoRA & 72.416/27.584 & 1189698 & 165.3416/246.2 & 5.2586 \\

\midrule

\multirow{2}{*}{DS-3}
& Base  & 52.825/47.175 & 1155104 & 124.0713/162.2 & 6.8952 \\
& QLoRA & 91.158/8.842  & 1166533 & 151.5415/209.8 & 5.6861 \\

\midrule

\multirow{2}{*}{DS-4}
& Base  & 55.536/44.464 & 1163749 & 209.7215/345.2 & 4.1181 \\
& QLoRA & 75.147/24.853 & 1173492 & 266.5912/460.6 & 3.2679 \\

\bottomrule
\end{tabular*}
\end{table}

Across all profiles, the blockchain layer records the resolved action, role assignment, lifecycle state, and execution metadata without altering the fraud decision. The reported average gas values range from 1,154,200 to 1,189,698 gas per transaction across the eight execution profiles. The on-chain AI-decision record contains the request identifier, model type, decision source, supervisor identifier, agreement state, resolved decision, confidence value, model and data hashes, and execution gas and latency metadata. Configuration-specific resolver-reason strings are retained off-chain and are not passed to the smart-contract decision record. The remaining variation in average gas reflects configuration-dependent lifecycle execution, including whether an additional \texttt{resolveReview} transaction is required, together with the values written by the executed contract path. The gas measurements should therefore be interpreted as observed execution overhead for the complete lifecycle profiles rather than as a fixed cost of storing an individual fraud decision.

Table~\ref{tab:blockchain_robustness} reports the offline artefact-reconstruction, payload-tampering, lifecycle-integrity, and caller-authorisation tests. The stored data and model digests are reconstructed exactly across all 40 canonical Phase-4 artefacts, with no cross-consensus digest or final-action mismatches. Digest recomputation also detects every tested nonzero payload modification across the eight reference datasets.

\begin{table*}[!t]
\centering
\caption{Blockchain robustness, lifecycle-integrity, and caller-authorisation assessment.}
\label{tab:blockchain_robustness}
\scriptsize
\setlength{\tabcolsep}{3pt}
\renewcommand{\arraystretch}{1.12}
\begin{tabularx}{\textwidth}{@{}L{2.25cm} L{2.7cm} X L{1.45cm} L{1.75cm} X@{}}
\toprule
\textbf{Assessment} &
\textbf{Scope} &
\textbf{Test Outcome} &
\textbf{Response} &
\textbf{State Effect} &
\textbf{Supported Conclusion} \\
\midrule

Digest reconstruction &
40 canonical Phase-4 artefacts &
Data and model digests reconstructed exactly; zero cross-consensus digest or final-action mismatches. &
Not applicable &
Not applicable &
Retained artefacts are internally consistent and hash-verifiable. \\

Offline payload tampering &
Eight System--Variant reference datasets &
Every tested nonzero payload modification was detected by at least one recomputed digest. &
Offline test &
Not applicable &
Post-export payload modification is detectable through digest recomputation. \\

Identical initial-decision resubmission &
\texttt{recordAIDecision} before finalisation &
Accepted and decision events were emitted again. &
Accepted &
No request-state change &
The initial-decision endpoint is not idempotent at the event-log level. \\

Conflicting initial-decision resubmission &
\texttt{recordAIDecision} before finalisation &
Accepted and the stored approval and fraud states changed. &
Accepted &
Request state changed &
Pre-finalisation initial decisions do not have single-write immutability. \\

Review and finalisation replay &
\texttt{resolveReview} and \texttt{finalizeRequest} &
Identical and conflicting review resubmissions, together with repeated finalisation, reverted. &
Rejected &
No request-state change &
Supervisor review resolution and finalisation are single-use in the tested paths. \\

Post-finalisation mutation &
\texttt{recordAIDecision} and \texttt{resolveReview} &
Conflicting decision and review updates reverted after successful finalisation. &
Rejected &
No request-state change &
Finalisation creates a terminal integrity boundary for the tested decision and review paths. \\

\midrule

unauthorised state-changing calls &
\texttt{recordAIDecision} and \texttt{resolveReview} on fresh local-chain requests &
Calls from an unauthorised account reverted with receipt status zero and produced no request-state change, whereas the corresponding authorised-validator control calls succeeded. &
Rejected; authorised controls accepted &
No state change for unauthorised calls &
Actor-based authorisation is enforced for the two tested state-changing endpoints under the evaluated local deployment. \\

\bottomrule
\end{tabularx}

\vspace{1mm}
\begin{minipage}{\textwidth}
\footnotesize\textit{Note:} Application-level duplicate submissions were issued as new Ethereum transactions with new account nonces. Replay of an identical signed raw transaction was not tested. Historical Phase-4 integrity was assessed from the retained artefacts because the original local Ganache state was not retained after artefact export; the live tests used fresh deployments of the retained ABI and bytecode.
\end{minipage}
\end{table*}

The live-chain tests reveal lifecycle-dependent protection. Before finalisation, an identical AI-decision resubmission is accepted and re-emits events without changing the request state, while a conflicting resubmission is accepted and changes the stored approval and fraud states. Repeated review resolution and repeated finalisation are rejected. After successful finalisation, conflicting calls to both \texttt{recordAIDecision} and \texttt{resolveReview} revert without changing the stored state. The tested contract enforces single-use review resolution and post-finalisation integrity, but does not provide universal pre-finalisation single-write protection. In particular, the live-chain tests show that a conflicting authorised resubmission to \texttt{recordAIDecision} can overwrite the stored decision state before finalisation. Production implementations should therefore enforce one-time initial-decision recording before permitting subsequent review or finalisation operations.

Unauthorised calls to \texttt{recordAIDecision} and \texttt{resolveReview} revert with receipt status zero and cause no request-state change, whereas the corresponding authorised-validator controls succeed. This supports actor-based authorisation for these two endpoints in the tested local deployment, but not resilience to stolen keys, compromised authorised actors, consensus-layer attacks, denial-of-service, or untested endpoints.

\section{Discussion}
\label{sec:discussion}

The results show that DS systems must be evaluated by their end-to-end operating profile rather than by the presence of a second model. The key issue is whether supervisory intervention recovers enough fraud to justify the resulting legitimate-user burden and \textsc{Review} workload.

\subsection{Are Supervisors Always Beneficial?}

On the complete matched population, no decision-changing Supervisor improves F1 or total errors relative to its corresponding primary-only stage. DS-1 Base adds 44 true positives but 29,182 false positives, while DS-3 Base adds 94 true positives but 38,542 false positives. The recall gains arise from broad escalation and do not indicate a better overall error profile. DS-1 Base is particularly unsuitable as an operational setting under the evaluated replay: supervision reduces accuracy to 55.13\%, raises FPR to 47.18\%, and blocks 49.78\% of all requests. Such a profile would impose an excessive legitimate-request burden and should not be treated as a viable deployment operating point.

Lower intervention rates produce less disruptive outcomes, but even the more selective pipelines do not outperform their primary-only baselines on F1 or total errors. Because no operator-specific cost model is estimated, these findings characterise the measured fraud-recovery and legitimate-user-blocking trade-offs; they do not establish economic net benefit.

\subsection{Why Base LLM Supervisors Can Be Over-Conservative}

The Base LLM Supervisors produce broad escalation under the baseline-derived thresholds and binary resolver. In DS-3 Base, the FedAvg primary stage has an F1 score of 71.64\% and a FPR of 3.88\%; after supervision, these become 18.94\% and 44.45\%, respectively.

Because both reliable \textsc{Review} and \textsc{Block} outcomes become final blocks, broad supervisory signals directly amplify legitimate false positives. This does not show that Base LLMs are inherently unsuitable for supervision; it indicates that their score distributions require stricter candidacy, threshold, and resolver control in an enforcement-oriented pipeline.

\subsection{Base--QLoRA Differences When the LLM Is the Supervisor}

In DS-1 and DS-3, the evaluated QLoRA Supervisor is markedly more selective than the Base Supervisor within the corresponding role-aware configurations. In DS-3, QLoRA produces 304 interventions, adding four true positives and 300 false positives, compared with 38,636 interventions, 94 true positives, and 38,542 false positives for Base. Its final profile remains much closer to the FedAvg primary stage, with 95.87\% accuracy, 97.08\% recall, 70.12\% F1, and a FPR of 4.20\%.

This difference should not be attributed causally to QLoRA adaptation alone because the variants retain their own realised artefacts, score distributions, and baseline-derived thresholds. DS-3 QLoRA also does not improve on the FedAvg primary stage overall: recall rises by 0.08 percentage points and soft recall by 0.80 percentage points, but F1 falls by 1.52 percentage points, FPR rises by 0.32 percentage points, and total errors increase by 296. The evidence supports a narrower conclusion: the evaluated QLoRA Supervisor is more controlled than the evaluated Base Supervisor under the tested pipeline.

\subsection{Role Assignment Matters}

The directional comparisons reflect complete implemented configurations and should not be interpreted as isolated causal role swaps. Each direction retains its role-specific candidacy, evidence-validity, threshold, and resolution conditions. The score-transfer audit confirms that retained component scores remain identical on pair-common support within the $1\times10^{-12}$ tolerance, but it does not equalise score availability or the surrounding operating contracts.

For the centralised-ML--LLM pair, both role directions produce decision-changing supervision, but their relative performance depends on the LLM variant. DS-2 has the stronger Base profile, whereas DS-1 has the stronger QLoRA profile. In DS-2, centralised-ML supervision increases fraud recall but introduces enough legitimate false positives to reduce F1 and increase total errors relative to the corresponding LLM-primary stage. For the FedAvg--LLM pair, DS-3 QLoRA has the highest F1 and lowest FPR among the supervised resolved configurations, although it remains weaker than its primary FedAvg stage. The reverse DS-4 QLoRA direction produces substantially lower F1 and higher FPR. Role-aware design must assess the complete pipeline. Strong performance as a Decider does not imply value as a Supervisor, and directional differences cannot be attributed to role labels alone.

\subsection{Dependability, Review Workload, and Failure Handling}

Bounded supervision does not guarantee better final decisions. Across the paired-bootstrap and local threshold-sensitivity analyses, every intervention-producing pipeline retains lower F1 and higher FPR than its corresponding primary-only stage. A direct-block-only resolver reduces the false-positive burden created by review-mediated escalation, but does not overturn the primary-stage comparison.

Traffic-mix reweighting shows that these trade-offs depend on fraud prevalence and severity composition. Supervision reduces total errors only for DS-4 Base in two extreme scenarios combining 20\% fraud prevalence with a 75\% or 100\% soft-fraud share. DS-2 Base achieves higher F1 in 14 of the 36 scenarios but does not reduce total errors in any scenario, while DS-2 QLoRA shows neither higher F1 nor fewer errors. Higher F1 can therefore occur without a reduction in incorrect decisions, and deployment claims must be conditioned on the expected traffic mix and the relative consequences of false positives and false negatives.

Keeping \textsc{Review} unresolved transfers immediate blocking burden to a finite queue; it does not remove that burden. The smaller DS-1 QLoRA and DS-3 QLoRA queues fit within a capacity of 100 requests per 100,000, whereas the larger DS-2 and DS-4 queues remain dominated by legitimate requests. For DS-2 Base and QLoRA, the first 100 ranked requests capture only three of 19 and one of 17 queued fraud cases, respectively. In DS-4 QLoRA, the first 100 ranked requests contain no fraud, while the first 1,000 capture seven of eight queued fraud cases but also include 993 legitimate requests. These results describe queue composition and oracle bounds, not reviewer accuracy, delay, cost, or final adjudication.

The Pareto and normalised cost-ratio analyses reinforce this trade-off. The DS-3 QLoRA final output recovers four additional fraud cases relative to the primary FedAvg output but introduces 300 additional legitimate blocks. With false-positive cost normalised to one and \textsc{Review} cost set to zero, it becomes lower-cost only when $\lambda_{\mathrm{FN}}>75$. Recent industry evidence illustrates the economic scale of telecommunications fraud: TNS, drawing on the latest CFCA Fraud Loss Survey, reports that global telecommunications fraud losses increased from \$38.95 billion in 2023 to \$41.82 billion in 2025 \cite{tns2026fraud}. However, such aggregate industry-loss estimates do not provide compatible per-request costs for a missed fraud case and an incorrectly blocked legitimate request, and therefore cannot establish that a 75:1 cost ratio is realistic for a particular operator. The value $\lambda_{\mathrm{FN}}>75$ should consequently be interpreted only as the measured break-even boundary of the normalised sensitivity analysis, not as an empirically supported telecom cost ratio.

Complete Supervisor-evidence loss safely reproduces the primary-only decisions, but also removes incremental fraud recovery. Systematic score bias and K2-O targeted manipulation show that score provenance and threshold governance are equally important because eligible requests near intervention boundaries can be moved into or out of escalation. The K2-O result remains an oracle worst-case stress test and does not establish threshold-agnostic adversarial robustness.

Primary-LLM failure is different because the primary action itself is unavailable. The fail-open, fail-closed, unresolved-\textsc{Review}, and DS-2 centralised-ML action-only takeover analyses are hypothetical policy sensitivities and do not represent implemented failover behaviour. Production use would require an authenticated failure signal, explicit authority transfer, complete secondary-evidence coverage, bounded \textsc{Review} capacity, and a tested recovery transition.

\subsection{Practical Implications for Network-Fraud Decision Management}

Deployment decisions should not be based on recall alone. FPR, block rate, false positives added, and FP per TP gain are needed to show whether additional fraud recovery is achieved through disproportionate blocking. Selective intervention is preferable to broad escalation, but even a low intervention rate can increase total errors. Each supervised pipeline should be compared with its own primary-only stage and assessed using operator-specific weights for missed fraud, legitimate-user blocking, \textsc{Review} workload, and service disruption. Role assignment is likewise an operational design choice and should be assessed at the complete-pipeline level rather than attributed to model families or role labels in isolation.

Blockchain contributes auditability and lifecycle enforcement, not predictive quality. The tested contract records role and decision metadata, supports digest reconstruction, enforces authorisation for \texttt{recordAIDecision} and \texttt{resolveReview}, permits only one review resolution, and protects the tested states after finalisation. It does not provide universal pre-finalisation single-write protection because a conflicting initial-decision submission can alter the stored state before finalisation. Production deployment requires explicit duplicate or idempotency controls, broader endpoint testing, and stronger key-management protection.

\section{Threats to Validity and Limitations}
\label{sec:threats}

The principal interpretation boundaries concern the controlled synthetic evaluation setting, the predefined role and policy configurations, the evaluated LLM variants and failure-handling assumptions, and the scope of the blockchain implementation. These boundaries define the conditions under which the reported DS comparisons and lifecycle results should be interpreted.

\subsection{Dataset, Statistical Scope, and Generalisation}

The study reuses the controlled synthetic network-request corpus, labels, and component artefacts from the companion study. This common data representation and preprocessing support consistent comparison across DS-1--DS-4 and controlled analysis of hard/soft fraud, traffic composition, and supervisory intervention. The evaluation is nevertheless an offline synthetic replay rather than validation on live operator traffic. Soft fraud is design-specific and denotes labelled fraud that remains within the admissible synthetic envelope without activating the hard-fraud gate; the matched replay contains 500 such requests.

The traffic-mix analysis varies fraud prevalence and the soft-fraud share while preserving the measured class-conditional behaviour, retained actions, thresholds, and within-class distributions. It therefore evaluates sensitivity to class composition rather than broader forms of distributional or behavioural shift. Similarly, the paired stratified bootstrap intervals quantify request-sampling uncertainty within the matched replay rather than uncertainty arising from alternative corpus generation, retraining, or deployment conditions.

The findings should therefore be interpreted as evidence for the evaluated artefacts, thresholds, role configurations, and policies under the tested replay conditions. Validation on independently generated or operational datasets, repeated training runs, and prospective temporal evaluation would be needed to assess transfer to other deployment environments.

\subsection{Role, Policy, and Utility Scope}

The study evaluates four predefined DS pipelines in both implemented role directions. The comparisons therefore characterise the complete operating profiles produced by their role-specific candidacy, evidence, thresholds, and resolution rules rather than an isolated causal effect of exchanging Decider and Supervisor labels.

The reported supervisory effects are conditional on the implemented no-downgrade binary resolver and the predefined baseline-derived thresholds. The direct-block-only and three-way \textsc{Review} analyses demonstrate how alternative resolution policies redistribute burden between immediate blocking and unresolved workload. The validation-only conditional-calibration analysis additionally tests whether transferring standalone thresholds explains the main results; although some selected operating points improve individual metrics, none reduces deployment total errors below the corresponding primary-only stage. The reported results therefore describe the tested policy and threshold configurations rather than all possible supervisory operating policies.

The normalised cost analysis complements these comparisons by varying dimensionless relative weights assigned to false negatives, false positives, and selected \textsc{Review} cases. These weights are analytical rather than operator-derived and are used to identify break-even regions, not to claim a universal economic optimum. Operator-specific deployment decisions would require empirical loss, service-level, and review-cost information.

\subsection{LLM Variant and Failure-Handling Scope}

The experiments compare one evaluated Base LLM variant with one QLoRA-adapted variant under the retained artefacts, baseline-derived thresholds, and role-aware decision policies. In DS-1 and DS-3, the QLoRA Supervisor is more selective and less disruptive than Base, whereas in DS-2 and DS-4 the Base--QLoRA differences already arise at the primary stage. These results therefore characterise the evaluated variants and complete pipelines rather than an isolated causal effect of QLoRA adaptation.

The DS-2 and DS-4 lifecycle artefacts do not implement a run-time authority-transfer mechanism for unavailable primary-LLM output. The reported fail-open, fail-closed, unresolved-\textsc{Review}, and DS-2 centralised-ML takeover results are therefore post hoc policy sensitivities rather than implemented failover behaviour. In particular, DS-4 does not provide FedAvg evidence for the original LLM-block population, so a complete FedAvg takeover cannot be evaluated. 

\subsection{Blockchain Execution and Integrity Scope}

The blockchain layer is evaluated as a local audit and lifecycle-enforcement backend rather than a production-scale blockchain benchmark. At the observed average execution costs of 1,154,200--1,189,698 gas per transaction, individually recording 100,000 requests corresponds to approximately 115.42--118.97 billion gas, or about 3,847--3,966 full-block equivalents when 30 million gas per block is used solely as an illustrative capacity benchmark. A production-oriented architecture could instead retain request-level records off-chain and periodically anchor batch commitments, such as Merkle roots, on-chain; the illustrative anchoring counts reported in this study describe batching frequency only, because Merkle construction, proof verification, anchoring cost, latency, and storage savings were not implemented or measured.

The original Ganache state used for Phase-4 execution was not retained, so the integrity analysis combines digest reconstruction of the retained artefacts with lifecycle tests on fresh deployments of the retained contract ABI and bytecode. The duplicate-submission tests assess application-level handling through new Ethereum transactions rather than replay of an identical signed transaction. Before finalisation, identical and conflicting initial-decision submissions can be accepted, and a conflicting submission can alter the stored request state. By contrast, repeated review resolution, repeated finalisation, and the tested post-finalisation mutations are rejected. The contract therefore does not provide universal pre-finalisation single-write protection or general replay resistance.

The authorisation tests cover \codefield{recordAIDecision} and \codefield{resolveReview} on fresh local deployments. Unauthorised calls to these endpoints revert without changing request state, while the corresponding authorised controls succeed. These results establish the tested lifecycle and authorisation properties within the evaluated local implementation; broader endpoint and adversary coverage remains outside the tested boundary.

\section{Conclusion and Future Work}
\label{sec:conclusion}

This paper presented a role-aware DS framework for network-fraud decision management with blockchain auditability. Four predefined configurations assigned centralised ML, a FedAvg-trained federated meta-model, and LLM-family components to Decider and Supervisor roles under a shared conservative resolution policy. The framework formalised bounded supervisory intervention through candidacy, evidence-validity, no-downgrade, threshold, and resolution constraints and evaluated complete pipelines using predictive, intervention-burden, unresolved-\textsc{Review}, dependability, and blockchain-integrity measures.

The results show that supervision is not inherently beneficial. Base LLM Supervisors recover limited additional fraud in DS-1 and DS-3 while introducing substantial false-positive burden. DS-3 QLoRA has the highest F1 and lowest FPR among the supervised resolved configurations, but remains weaker than its primary FedAvg stage in F1, legitimate-user preservation, and total errors. Directional differences between DS-1 and DS-2 and between DS-3 and DS-4 characterise complete role-specific configurations and do not isolate the causal effect of exchanging the role labels.

The operational analyses further show that these conclusions depend on traffic composition, resolver design, \textsc{Review} capacity, and relative error costs. Across 36 exactly reweighted traffic mixtures, supervision reduces total errors only for DS-4 Base in the two scenarios combining 20\% fraud prevalence with a 75\% or 100\% soft-fraud share. Leaving \textsc{Review} unresolved reduces immediate automated blocking but transfers the burden into a finite queue. The DS-1 QLoRA and DS-3 QLoRA queues fit within a capacity of 100 requests per 100,000, whereas the larger DS-2 and DS-4 queues are overwhelmingly legitimate and require substantially greater capacity for complete processing. Under the normalised cost model, the DS-3 QLoRA final output becomes lower-cost than the primary FedAvg output only when $\lambda_{\mathrm{FN}}>75$; this is an analytical break-even boundary, not an empirically estimated operator cost.

Complete Supervisor-evidence loss safely reproduces the primary decisions, but systematic score bias and K2-O targeted threshold-crossing manipulation can sharply alter intervention outcomes. The queue-capacity and K2-O results are analytical oracle bounds; they do not represent measured reviewer performance or establish threshold-agnostic adversarial robustness. Primary-LLM failure likewise remains dependent on hypothetical fail-open, fail-closed, unresolved-\textsc{Review}, or takeover policies because no run-time authority-transfer contract is implemented.

The blockchain layer provides auditability and lifecycle enforcement without contributing predictive performance. Digest reconstruction confirms the internal consistency of the retained artefacts, and recomputation detects every tested nonzero payload modification. Repeated review resolution, repeated finalisation, and the tested post-finalisation mutations are rejected. Unauthorised calls to \texttt{recordAIDecision} and \texttt{resolveReview} also revert without changing request state, while the corresponding authorised-validator controls succeed. However, identical and conflicting initial-decision submissions can be accepted before finalisation, including a conflicting submission that changes the stored request state. The supported protections are specific to the tested endpoints, callers, lifecycle states, and local deployment.

Future work will implement the unresolved-\textsc{Review} pathway as a live bounded service with authenticated queue admission, explicit capacity and overflow rules, measurable delay and resource use, and measured post-review outcomes. It will also examine broader LLM architectures, prompts, adaptation and calibration procedures, independently generated and operational datasets, repeated training runs, temporal drift, and threshold-agnostic adversarial strategies. Production-oriented failure handling will require authenticated primary-failure signals, complete secondary-evidence coverage, explicit authority transfer, bounded unresolved-request handling, and tested recovery to normal operation. The smart-contract lifecycle will additionally require pre-finalisation single-write or idempotency controls, broader endpoint and role-governance testing, validator-key protection, compromised-authorised-actor scenarios, alternative consensus settings, and evaluation under production network conditions.

\section*{CRediT authorship contribution statement}

\textbf{Saviz Changizi:} Conceptualization, Methodology, Software, Validation, Formal analysis, Investigation, Data curation, Writing -- original draft, Writing -- review \& editing, Visualization.

\textbf{Nasibeh Mohammadzadeh:} Writing -- original draft, Writing -- review \& editing.

\textbf{Mohammad Shojafar:} Supervision, Writing -- review \& editing.

\textbf{Rahim Tafazolli:} Supervision.

\section*{Data and Code Availability}

The reproducibility materials are organised across two institutional GitLab repositories. The \href{https://gitlab.surrey.ac.uk/sc02731/blockchain-linked-auditable-decision-management-for-telecom-spectrum-access-fraud-control-in-5g-6g-inspired-heterogeneous-telecom-iot-environments}{companion GitLab repository} contains the shared dataset-construction pipeline, standalone M1--M3 risk-source artefacts, and blockchain-linked audit components reused in this study.

The \href{https://gitlab.surrey.ac.uk/sc02731/when-does-ai-supervision-help-a-role-aware-study-of-network-fraud-decision-management-with-blockchain-auditability}{present-paper GitLab repository} contains the role-aware DS evaluation notebooks, DS-1--DS-4 result files, generated figures and LaTeX tables, intervention and directional-comparison analyses, matched score-transfer audits, conditional-validation calibration analyses, paired-bootstrap uncertainty and policy-sensitivity outputs, component-failure and hypothetical primary-LLM failure-resolution analyses, and blockchain audit, digest-reconstruction, tamper-detection, and lifecycle-test artefacts. It also includes the operational-validation materials for population alignment, probability provenance, traffic-mix reweighting, finite \textsc{Review} capacity, label-free queue ranking, overflow policies, Pareto and normalised cost-ratio analyses, oracle bounds, validation manifests, the figure registry, and the hashed output inventory.

\section*{Funding}

This work was supported by the Engineering and Physical Sciences Research Council (EPSRC) [grant number EP/Y035534/1].

\bibliographystyle{elsarticle-harv}
\bibliography{references}

@article{ali2023xai,
  author = {Ali, S. and Abuhmed, T. and El-Sappagh, S. and Muhammad, K. and Alonso-Moral, J. M. and Confalonieri, R. and Guidotti, R. and Del Ser, J. and Diaz-Rodriguez, N. and Herrera, F.},
  title = {Explainable Artificial Intelligence ({XAI}): What we know and what is left to attain Trustworthy Artificial Intelligence},
  journal = {Information Fusion},
  volume = {99},
  pages = {101805},
  year = {2023},
  doi = {10.1016/j.inffus.2023.101805}
}

@article{shorinwa2025uncertainty,
  author = {Shorinwa, O. and Mei, Z. and Lidard, J. and Ren, A. Z. and Majumdar, A.},
  title = {A Survey on Uncertainty Quantification of Large Language Models: Taxonomy, Open Research Challenges, and Future Directions},
  journal = {ACM Computing Surveys},
  volume = {58},
  number = {3},
  pages = {63},
  year = {2025},
  doi = {10.1145/3744238}
}

@article{bach2024humanai,
  author = {Bach, T. A. and Kristiansen, J. K. and Babic, A. and Jacovi, A.},
  title = {Unpacking Human-{AI} Interaction in Safety-Critical Industries: A Systematic Literature Review},
  journal = {IEEE Access},
  volume = {12},
  pages = {106385--106414},
  year = {2024},
  doi = {10.1109/ACCESS.2024.3437190}
}

@article{vaccaro2024humanai,
  author = {Vaccaro, M. and Almaatouq, A. and Malone, T. W.},
  title = {When combinations of humans and {AI} are useful: A systematic review and meta-analysis},
  journal = {Nature Human Behaviour},
  volume = {8},
  pages = {2293--2303},
  year = {2024},
  doi = {10.1038/s41562-024-02024-1}
}

@article{liu2024gap,
  author = {Liu, B. and Li, G. and Zhang, H. and Jin, Y. and Wang, Z. and Shao, D.},
  title = {The Gap Between Trustworthy {AI} Research and Trustworthy Software Research: A Tertiary Study},
  journal = {ACM Computing Surveys},
  volume = {57},
  number = {3},
  pages = {52},
  year = {2024},
  doi = {10.1145/3694964}
}

@article{kemmerzell2025trust,
  author = {Kemmerzell, N. and Schreiner, A. and Khalid, H. and Schalk, M. and Bordoli, L.},
  title = {Towards a Better Understanding of Evaluating Trustworthiness in {AI} Systems},
  journal = {ACM Computing Surveys},
  volume = {57},
  number = {9},
  pages = {218},
  year = {2025},
  doi = {10.1145/3721976}
}

@article{wu2024fraud,
  author = {Wu, J. and Hu, R. and Li, D. and Ren, L. and Huang, Z. and Zang, Y.},
  title = {Beyond the Individual: An Improved Telecom Fraud Detection Approach Based on Latent Synergy Graph Learning},
  journal = {Neural Networks},
  volume = {169},
  pages = {20--31},
  year = {2024},
  doi = {10.1016/j.neunet.2023.10.019}
}

@article{chen2026blockchainids,
  author = {Chen, B. and Wang, Y. and Santos, L. and Yang, X. and Liu, C. and Im, S.-K.},
  title = {A Four-Paradigm Taxonomy and Systematic Survey of Blockchain-Enabled Intrusion Detection Systems for {IoT} and {IIoT}},
  journal = {IEEE Internet of Things Journal},
  volume = {13},
  number = {10},
  pages = {20312--20338},
  year = {2026},
  doi = {10.1109/JIOT.2026.3662349}
}

@article{wu2026blockchainfl,
  author = {Wu, Y. and Cheng, H. and Liu, X. and Xu, Y. and Chen, F. and Zhang, X. and Wang, M.},
  title = {Blockchain-Based Secure Federated Learning With Improved Consensus Protocol and Personalized Differential Privacy},
  journal = {IEEE Transactions on Dependable and Secure Computing},
  volume = {23},
  number = {2},
  pages = {2412--2429},
  year = {2026},
  doi = {10.1109/TDSC.2025.3626847}
}

@article{singh2024flids,
  author = {Singh, G. and Sood, K. and Rajalakshmi, P. and Nguyen, D. D. N. and Xiang, Y.},
  title = {Evaluating Federated Learning-Based Intrusion Detection Scheme for Next Generation Networks},
  journal = {IEEE Transactions on Network and Service Management},
  volume = {21},
  number = {4},
  pages = {4816--4829},
  year = {2024},
  doi = {10.1109/TNSM.2024.3385385}
}

@article{han2026fedehp,
  author = {Han, S. and Pan, J. and Zhao, S. and Ren, S. and Wang, Z. and He, S. and Ren, K. and Qin, Z. and Chen, X.},
  title = {{Fed-EHP}: Efficient and Heterogeneous Privacy-Preserving Personalized Federated Learning},
  journal = {IEEE Transactions on Dependable and Secure Computing},
  volume = {23},
  number = {2},
  pages = {3206--3221},
  year = {2026},
  doi = {10.1109/TDSC.2025.3634446}
}

@article{hu2024gatcobo,
  author = {Hu, X. and Chen, H. and Zhang, J. and Liu, S. and Li, X. and Wang, Y. and Xue, X.},
  title = {{GAT-COBO}: Cost-Sensitive Graph Neural Network for Telecom Fraud Detection},
  journal = {IEEE Transactions on Big Data},
  volume = {10},
  number = {4},
  pages = {528--542},
  year = {2024},
  doi = {10.1109/TBDATA.2024.3352978}
}

@article{li2026privacyfl,
  author = {Li, Z. and Kong, Y. and Jiang, C.},
  title = {Privacy-Aware Transaction Fraud Detection With Low Communication Costs Under Joint Federated Learning},
  journal = {IEEE Transactions on Dependable and Secure Computing},
  volume = {23},
  number = {3},
  pages = {4511--4525},
  year = {2026},
  doi = {10.1109/TDSC.2025.3647476}
}

@article{zhang2025verifiablefl,
  author = {Zhang, C. and Weng, J. and Weng, J. and Zhong, Y. and Liu, J.-N. and Deng, C.},
  title = {Robust and Secure Federated Learning With Verifiable Differential Privacy},
  journal = {IEEE Transactions on Dependable and Secure Computing},
  volume = {22},
  number = {5},
  pages = {5713--5729},
  year = {2025},
  doi = {10.1109/TDSC.2025.3574745}
}

@article{kamran2026blockchainfl,
  author = {Kamran, M. and Akhtar, S. M. and Gilani, A. and Alhashmi, A. A. and Kanwal, S. and Darem, A. A. and Alofairi, A. A.},
  title = {A Blockchain-Assisted Secure Federated Learning Architecture for Intrusion Detection in Internet of Things Networks},
  journal = {Scientific Reports},
  year = {2026},
  doi = {10.1038/s41598-026-53053-x}
}

@article{alshammari2026securetrust,
  author = {Alshammari, N. S. and Mishra, S. and Rathi, M. and Goel, N. and Tahzib, S.},
  title = {{SecureTrust-FL}: Trust-Aware Privacy-Preserving Federated Learning for Network Intrusion Detection},
  journal = {Scientific Reports},
  year = {2026},
  doi = {10.1038/s41598-026-58383-4}
}

@article{zhang2026contrastivefl,
  author = {Zhang, H. and Yu, Z. and Wang, G. and Xu, F. and Zhang, Y. and Hu, C. and Wang, X. and Yu, J.},
  title = {Toward Model-Contrastive Federated Learning With Lightweight Privacy Preservation and Poisoning Attack Detection},
  journal = {IEEE Transactions on Dependable and Secure Computing},
  volume = {23},
  number = {2},
  pages = {1830--1846},
  year = {2026},
  doi = {10.1109/TDSC.2025.3620529}
}

@inproceedings{changizi2025hpcc,
  author = {Changizi, S. and Mohammadzadeh, N. and Shojafar, M.},
  title = {Hybrid Telecom Fraud Detection with Machine Learning, Large Language Models, and Blockchain},
  booktitle = {Proceedings of the 2025 {IEEE} International Conference on High Performance Computing and Communications ({HPCC})},
  pages = {1100--1108},
  year = {2025},
  doi = {10.1109/HPCC67675.2025.00158}
}

@article{changizi2026blockchainlinked,
  author  = {Saviz Changizi and Nasibeh Mohammadzadeh and Mohammad Shojafar and Rahim Tafazolli},
  title = {Blockchain-Linked Auditable Decision Management for Telecom/{IoT} Fraud-Control Requests},
  journal = {{arXiv} preprint {arXiv}:2607.09259},
  year    = {2026},
  month   = jul,
  doi     = {10.48550/arXiv.2607.09259},
  url     = {https://doi.org/10.48550/arXiv.2607.09259}
}

@article{valdezsolis2026privacy,
  author = {Valdez-Solis, W. and Parra-Ullauri, J. and Simeonidou, D. and Kertesz, A.},
  title = {Privacy and Security for {6G} Networks via Fog/Edge Computing, Blockchain, and Federated Learning: A Survey and Taxonomy},
  journal = {Journal of Network and Computer Applications},
  volume = {252},
  pages = {104510},
  year = {2026},
  doi = {10.1016/j.jnca.2026.104510}
}

@article{belenguer2023gowfed,
  author = {Belenguer, A. and Pascual, J. A. and Navaridas, J.},
  title = {{G{\"o}wFed}: A Novel Federated Network Intrusion Detection System},
  journal = {Journal of Network and Computer Applications},
  volume = {217},
  pages = {103653},
  year = {2023},
  doi = {10.1016/j.jnca.2023.103653}
}

@article{racherla2024deepids,
  author = {Racherla, S. and Sripathi, P. and Faruqui, N. and Kabir, M. A. and Whaiduzzaman, M. and Shah, S. A.},
  title = {{Deep-IDS}: A Real-Time Intrusion Detector for {IoT} Nodes Using Deep Learning},
  journal = {IEEE Access},
  volume = {12},
  pages = {63584--63597},
  year = {2024},
  doi = {10.1109/ACCESS.2024.3396461}
}

@article{bozorgchenani2024intrusion,
  author = {Bozorgchenani, A. and Zarakovitis, C. C. and Chien, S. F. and Ni, Q. and Gouglidis, A. and Mallouli, W. and Lim, H. S.},
  title = {Intrusion Response Systems for the {5G} Networks and Beyond: A New Joint Security-vs-{QoS} Optimization Approach},
  journal = {IEEE Transactions on Network Science and Engineering},
  volume = {11},
  number = {3},
  pages = {3039--3052},
  year = {2024},
  doi = {10.1109/TNSE.2024.3358170}
}

@article{mhamdi2024securing,
  author = {Mhamdi, L. and Isa, M. M.},
  title = {Securing {SDN}: Hybrid Autoencoder-Random Forest for Intrusion Detection and Attack Mitigation},
  journal = {Journal of Network and Computer Applications},
  volume = {225},
  pages = {103868},
  year = {2024},
  doi = {10.1016/j.jnca.2024.103868}
}

@article{devipriya2025blockchain,
  author = {Devi Priya, V. S. and Sethuraman, S. C. and Khan, M. K.},
  title = {Blockchain-Based Deep Learning Models for Intrusion Detection in Industrial Control Systems: Frameworks and Open Issues},
  journal = {Journal of Network and Computer Applications},
  volume = {243},
  pages = {104286},
  year = {2025},
  doi = {10.1016/j.jnca.2025.104286}
}

@inproceedings{madras2018defer,
  author = {Madras, David and Pitassi, Toni and Zemel, Richard},
  title = {Predict Responsibly: Improving Fairness and Accuracy by Learning to Defer},
  booktitle = {Advances in Neural Information Processing Systems},
  volume = {31},
  year = {2018},
  url = {https://proceedings.neurips.cc/paper/2018/hash/09d37c08f7b129e96277388757530c72-Abstract.html}
}

@inproceedings{mozannar2020defer,
  author = {Mozannar, Hussein and Sontag, David},
  title = {Consistent Estimators for Learning to Defer to an Expert},
  booktitle = {Proceedings of the 37th International Conference on Machine Learning},
  series = {Proceedings of Machine Learning Research},
  volume = {119},
  pages = {7076--7087},
  year = {2020},
  url = {https://proceedings.mlr.press/v119/mozannar20b.html}
}

@inproceedings{charoenphakdee2021rejection,
  author = {Charoenphakdee, Nontawat and Cui, Zhenghang and Zhang, Yivan and Sugiyama, Masashi},
  title = {Classification with Rejection Based on Cost-sensitive Classification},
  booktitle = {Proceedings of the 38th International Conference on Machine Learning},
  series = {Proceedings of Machine Learning Research},
  volume = {139},
  pages = {1507--1517},
  year = {2021},
  url = {https://proceedings.mlr.press/v139/charoenphakdee21a.html}
}

@inproceedings{jitkrittum2023cascade,
  author = {Jitkrittum, Wittawat and Gupta, Neha and Menon, Aditya K. and Narasimhan, Harikrishna and Rawat, Ankit and Kumar, Sanjiv},
  title = {When Does Confidence-Based Cascade Deferral Suffice?},
  booktitle = {Advances in Neural Information Processing Systems},
  volume = {36},
  year = {2023},
  doi = {10.52202/075280-0431},
  url = {https://proceedings.neurips.cc/paper_files/paper/2023/hash/1f09e1ee5035a4c3fe38a5681cae5815-Abstract-Conference.html}
}

@misc{tns2026fraud,
  author = {{Transaction Network Services}},
  title = {The Telecom Fraud Landscape in 2026: How the Industry Is Fighting Back},
  year = {2026},
  month = jan,
  url = {https://tnsi.com/resource/com/the-telecom-fraud-landscape-in-2026-how-the-industry-is-fighting-back-blog/}
}

\end{document}